\documentclass[3p,times]{elsarticle}

\usepackage{ecrc}
\usepackage{graphicx}
\usepackage{dcolumn}
\usepackage{bm}
\usepackage[utf8]{inputenc}
\usepackage[T1]{fontenc}
\usepackage{mathptmx}
\usepackage{etoolbox}
\usepackage{color}
\usepackage{booktabs}
\usepackage{hyperref}
\usepackage{caption} 
\usepackage{subcaption}
\usepackage{enumitem}
\usepackage{amssymb}
\usepackage{amsmath}

\volume{00}

\firstpage{1}

\journalname{Computational Materials Science}

\runauth{Fern\'andez-Godino et al.}

\jid{procs}

\jnltitlelogo{Comp. Materials Science}

\CopyrightLine{2026}{Published by Elsevier Ltd.}

\usepackage[figuresright]{rotating}

\begin{document}

\begin{frontmatter}



\dochead{}

\title{Spatio-temporal, multi-field deep learning of shock propagation in meso-structured media}

\author[aff1]{M. Giselle Fern\'andez-Godino\corref{cor1}}
\ead{fernandez48@llnl.gov}
\cortext[cor1]{Corresponding author}
\author[aff1]{Meir H. Shachar}
\author[aff1]{Kevin Korner}
\author[aff1]{Jonathan L. Belof}
\author[aff1]{Mukul Kumar}
\author[aff1]{Jonathan Lind}
\author[aff1]{William J. Schill}

\address[aff1]{Lawrence Livermore National Laboratory, 7000 East Ave., Livermore, CA 94550, United States}

\begin{abstract}
Predicting the extreme hydrodynamic response of porous and architected lattice materials is a fundamental challenge in high energy density physics, where shock-induced pore collapse, baroclinic vorticity, and anomalous kinetic and thermodynamic states must be resolved across multiple scales. 
Traditional high-fidelity hydrocodes are computationally prohibitive for large-scale design exploration in applications like planetary defense and inertial confinement fusion. We present a multi-field spatio-temporal model (MSTM) designed to overcome the limitations of standard machine learning surrogates, which often fail to capture the sharp gradients and non-linear field couplings characteristic of shock propagation.
By training on high-fidelity, multiscale multiphysics data, MSTM simultaneously evolves seven coupled thermodynamic and kinetic fields, including pressure, temperature, density, and velocity, across complex material architectures. Our framework demonstrates strong predictive performance on held-out cases within the studied simulation setting, including anomalous responses such as counterintuitive post-shock density reductions and localized hotspot formation, with mean root mean squared errors as low as 1.4\%.
The model's multi-field formulation maintains low mass-conservation error and improved interface agreement over long autoregressive rollouts, outperforming single-field models by 94\% in structural fidelity. This framework enables a 1000$\times$ reduction in time to solution, providing a practical pathway for rapid analysis and optimization of energy dissipation and momentum transfer in meso-structured media.
\end{abstract}


\end{frontmatter}

\section{Introduction}

Shock waves traveling through porous or architected lattice materials drive phenomena across planetary defense and inertial confinement fusion, where controlling wave propagation is central to performance. When a shock traverses voided or architected lattice-structured media, pores collapse, localized heating develops, and anomalous Hugoniot responses emerge, including counterintuitive regimes in which higher shock pressures reduce post-shock density\cite{Pham2023,kline2024reducing}.

In planetary-defense studies, simulations performed by LLNL for the Double Asteroid Redirection Test (DART) show that an asteroid’s porosity strongly influences the momentum transferred by a kinetic impact: porous targets absorb more energy and compact instead of ejecting material, which reduces crater ejecta and alters the post-impact velocity change\cite{housen2018impacts,nichols2022porosity}.

For inertial-confinement fusion, LLNL’s polar-direct-drive wetted-foam targets use a thick liquid deuterium-tritium foam as both ablator and fuel; the foam lowers the in-flight aspect ratio and convergence ratio, suppresses Rayleigh-Taylor instabilities, improves hydrodynamic stability compared with traditional ice layers, and has been shown in HYDRA and xRAGE simulations to couple laser energy efficiently and achieve high thermonuclear yield\cite{dhakal2019effects}. LLNL has fabricated such targets by chemically growing ultralow-density aerogel foams inside spherical mandrels, and reducing the foam mass through additive manufacturing further improves fusion performance\cite{zhao2020dynamic}.

Composite metal foams made of hollow metal spheres embedded in a steel or aluminum matrix can absorb large fractions of dynamic loading energy while maintaining much lower weight than conventional solid metals. Studies of closed-cell aluminum foams show high energy absorption under dynamic loading, and graded-density designs further enhance performance\cite{zhao2020dynamic,parveez2022microstructure}. Composite metal foam combined with ceramic systems has also demonstrated the ability to dissipate 70–80\% of impact energy in laboratory tests\cite{garcia2014ballistic}.

In porous and architected lattice materials, shock interaction with densely distributed solid-void interfaces generates baroclinic vorticity in a manner analogous to the Richtmyer-Meshkov instability (RMI) at a single interface. While classical RMI describes jetting and mixing seeded by shock transit across a perturbed boundary, in meso-structured media the same mechanism is distributed across many interfaces, producing complex field evolution. Recent studies in high-energy-density physics have shown that RMI-driven jetting can be actively suppressed or enhanced through interface design and shock timing\cite{sterbentz2022design,kline2024reducing,sterbentz2024explosively,schill2024suppression}, highlighting the broader opportunity to control shock-induced instabilities by tailoring material architecture. This perspective motivates our focus on porous and architected lattice configurations, where interface-driven phenomena dominate the response.

Beyond these high-energy applications, shock-induced pore collapse also alters intrinsic material properties: it reduces ductility, strength, and toughness in metals and changes permeability and porosity in geological media, with implications for groundwater flow and resource recovery\cite{cheung2024data}. Understanding these processes requires resolving sharp gradients and rapidly evolving multi-material fields across multiple scales.

High-fidelity hydrocodes, including Eulerian and arbitrary Lagrangian-Eulerian formulations, remain a standard tool for shock simulations\cite{dobrev2012high,anderson2018high}. However, they typically employ millions of cells and very small time steps, making each simulation expensive and limiting brute-force design exploration or uncertainty quantification. This cost has motivated the development of surrogate models that retain physically relevant predictive fidelity while reducing inference time substantially.

Recent computational materials studies have used machine-learning surrogates to accelerate spatiotemporal prediction of evolving fields, including recurrent models for microstructure evolution\cite{Farizhandi2023,Qin2023} and machine-learning surrogates for high-rate fracture simulations\cite{Moore2018,fernandez2021accelerating,GarciaCardona2022}. Related computational studies have also examined shock response in porous systems, including shock-induced void collapse in nanoporous metals\cite{Wang2021}. Together, these studies motivate the present work on reduced-cost, multi-field prediction of shock response in porous and architected lattice materials.

We introduce a Multi-field Spatio-Temporal Model (MSTM) that autoregressively evolves the complete two-dimensional state of shocked porous and architected lattice materials. The model simultaneously predicts density, pressure, temperature, energy, two velocity components, and a material indicator using a hybrid convolutional-LSTM framework trained on high-fidelity hydrocode data. We train two separate instances of this same MSTM architecture, one on the porous dataset and one on the architected lattice dataset, and report results for each.

The main contributions of this work are:
\begin{enumerate}
\item We present a multi-field spatio-temporal surrogate for shock propagation in meso-structured materials that autoregressively predicts seven coupled fields, density, pressure, temperature, energy, material indicator, and two velocity components, from high-fidelity hydrocode data.
\item We evaluate the framework on both porous and architected lattice materials, where it achieves mean root mean squared errors of 1.4\% and 3.2\%, respectively, together with more than three orders of magnitude reduction in inference time relative to direct simulation.
\item We assess predictive performance on held-out cases within the studied simulation setting, including unseen porosities, lattice angles, and loading conditions. In the more challenging architected lattice case, MSTM maintains mass-averaged pressure, temperature, and density within about 5\% over the rollout.
\item We perform a controlled comparison against seven single-field spatio-temporal models under identical training and rollout settings, and show that joint multi-field prediction improves accuracy and structural fidelity.
\end{enumerate}

\section{Related Work}

A variety of surrogate families have been proposed to accelerate shock-driven simulations. Reduced-order models based on projection or manifold learning can speed up smooth flows, but linear bases smear moving discontinuities and nonlinear variants often require careful tuning and typically focus on a single physical field \cite{Mufti2024}. Physics-informed neural networks embed governing equations during training and solve parameterised PDEs, yet sharp shocks inhibit convergence and new scenarios usually require retraining \cite{Abbasi2025}. Neural operators map functions to functions: the Fourier Neural Operator learns mesh-invariant solution maps \cite{Li2020}, DeepONet extends operator learning to broader nonlinear problems \cite{Lu2021}, and the U-shaped Neural Operator improves multi-scale accuracy by combining spectral and U-Net features \cite{Rahman2023}. While fast, spectral representations tend to oversmooth localized shock fronts or introduce oscillations near discontinuities.

Graph-based simulators learn updates on meshes or particles; MeshGraphNets achieved resolution-robust rollouts for fluids and deformable solids \cite{Pfaff2021}, and graph surrogates have reproduced explosion-driven transients in complex 3D domains with large geometric generalization and substantial speedups \cite{Covoni2024}. These methods, however, have largely been developed for settings that differ from porous and architected lattice materials with coupled multi-field shock response.

Recent surrogate models for evolving materials fields include recurrent approaches for microstructure evolution, conv–recurrent surrogates for phase-field dynamics, and operator-learning surrogates for solidification problems \cite{Farizhandi2023,Qin2023,Wu2023,Ciesielski2025}. These studies address time-evolving materials systems, but they focus on governing physics and target outputs that differ from the present shock-dominated, multi-field hydrocode setting.

A complementary class of outcome-focused surrogates targets reduced descriptions of shock response rather than full spatio-temporal fields. Scarce-data models estimate Hugoniot curves across materials \cite{balakrishnan2023machine}; microstructure-to-hotspot mappings recover post-shock temperature fields \cite{Li2023}; and data-scarce surrogate models have also been developed for shock-induced pore-collapse dynamics in porous media \cite{cheung2024data}. More recently, latent-space dynamics frameworks have been introduced to track shock-induced pore-collapse interfaces, providing a compact representation of evolving morphology but without resolving the coupled multi-field state \cite{chung2025latent}. Bayesian neural fields supply spatio-temporal forecasts with uncertainty quantification, highlighting the importance of calibrated predictions for extreme-condition regimes \cite{saad2024scalable}. Machine-learning-driven design has also been pursued, for instance to reduce Richtmyer-Meshkov jet velocity in shaped charges \cite{kline2024reducing}; such inverse-design studies optimise a single metric rather than the coupled, time-resolved state evolution required for porous and architected lattice materials.

Collectively, these surrogate classes demonstrate the promise of data-driven modeling, but many existing approaches remain focused on single fields, reduced outcomes, fluid-only settings, or end-state predictions. Our MSTM framework addresses this gap by delivering time-resolved, multi-field predictions that evolve density, pressure, temperature, energy, material indicator, and velocity components together, while demonstrating predictive performance on held-out cases within the studied training distribution.

\section{Problem Description} \label{sec:level2}

\begin{figure}[!ht]
  \centering

  \begin{subfigure}[t]{0.295\textwidth}
    \vspace{0pt}                   
    \centering
    \includegraphics[width=\textwidth]{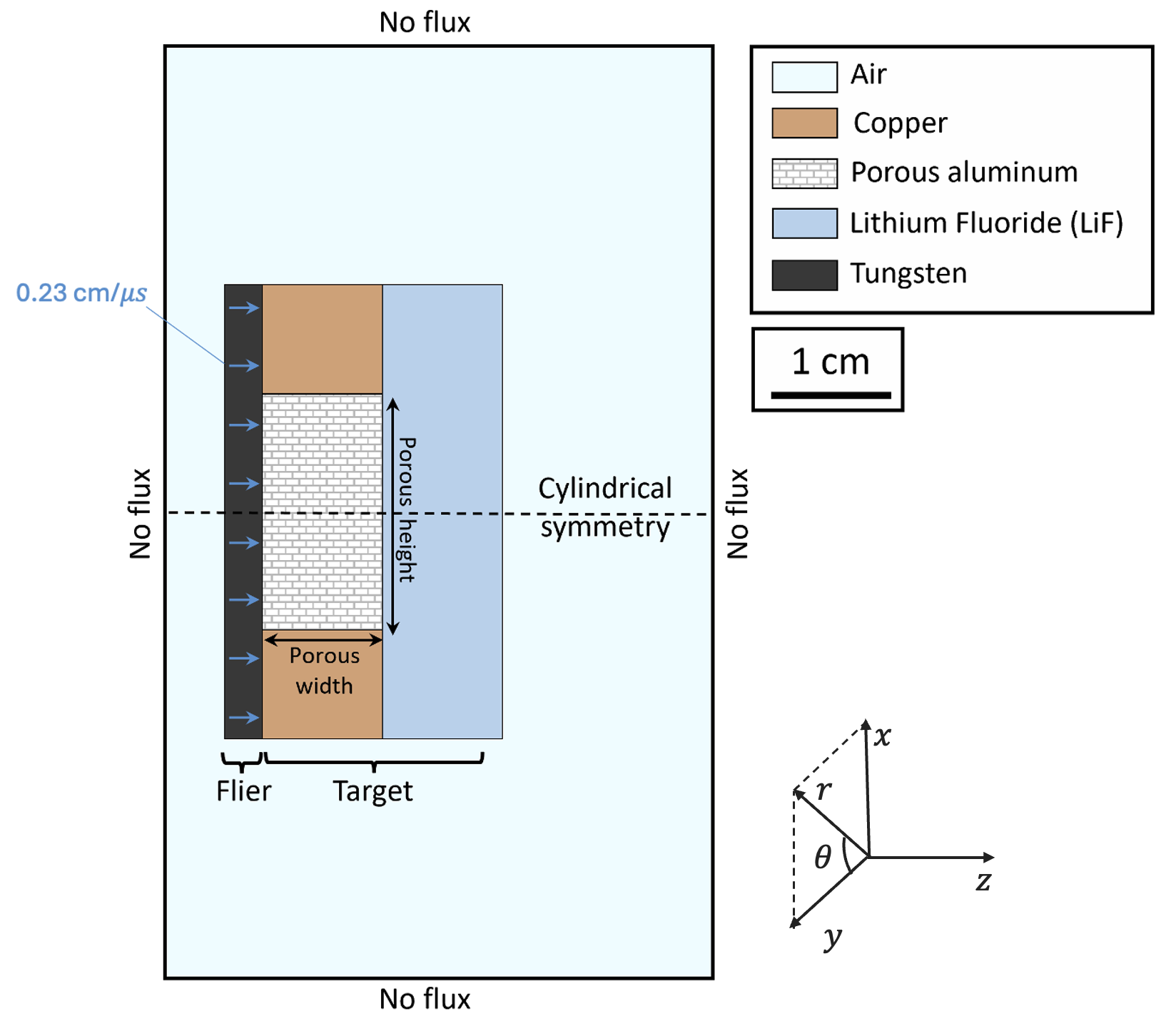}
    \caption{Porous material study: A two-dimensional cylindrically symmetric domain of 3.8 cm diameter and total axial extent \(1.3175~\mathrm{cm} + \textit{porous width}\), with symmetry about the central axis (dashed line). The setup includes a tungsten flier (0.3175 cm thick, impact velocity 0.23 cm/$\mu$s), a porous aluminum disc encased in a copper backer, and a lithium fluoride disc. The porous aluminum disc thickness, diameter, and homogenized porosity are varied over 0.2--1.0 cm, 0.05--3.8 cm, and 5\%--75\%, respectively.}
    \label{fig:setup:a}
  \end{subfigure}
  \hfill
  \begin{subfigure}[t]{0.50\textwidth}
    \vspace{0pt}                        
    \centering
    \includegraphics[width=\textwidth]{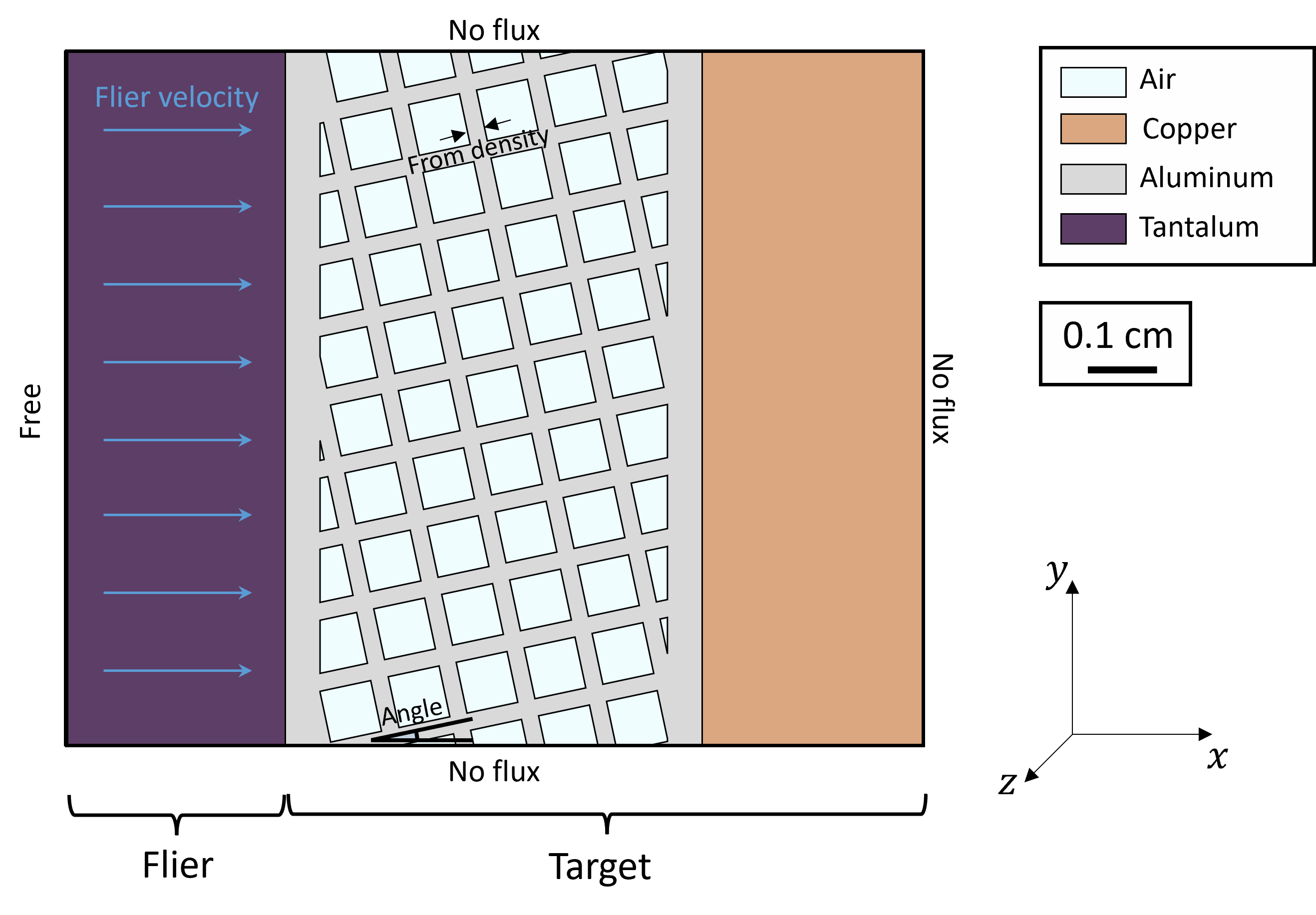}
    \caption{Architected lattice material study: A 0.3175 cm-thick tantalum flier impacts an architected aluminum lattice. The architected lattice is constructed by removing a grid of struts spaced at a 0.10 cm pitch and rotating the resulting pattern by a prescribed angle within the range [0$^{\circ}$, 45$^{\circ}$]. The strut thickness is determined from the porosity of the architected lattice, which is sampled from the range [10\%, 90\%]. Positioned downstream is a 0.3175 cm copper backer. In this setup, an additional control variable is the shock speed, which varies from 0.1 cm/$\mu$s to 0.4 cm/$\mu$s, and promotes controlled wave reflections and focusing.}
    \label{fig:setup:b}
  \end{subfigure}

  \caption{Simulation setup designed to study shock wave propagation through meso-structured media. The simulation output was used as the training, validation and test data set for the deep learning models.}
  \label{fig:setup}
\end{figure}

We simulate shock propagation through two classes of meso‑structured media: a porous aluminum disc and an architected aluminum lattice, each impacted by a metal flier. Figure~\ref{fig:setup} illustrates the geometries and denotes the key parameters.

\textbf{Porous disc.} A 0.3175 cm thick tungsten flier impacts a porous aluminum disc encased in a copper backer (Fig.~\ref{fig:setup:a}). The domain is cylindrically symmetric with diameter 3.8 cm and axial extent 1.3175 cm plus the porous aluminum disc thickness. The porous aluminum disc has variable \textit{thickness} (0.2–1.0 cm), \textit{diameter} (0.05–3.8 cm) and \textit{homogenized porosity} (5–75\%). The impact velocity is fixed at 0.23 cm/$\mu$s. This design allows systematic exploration of porosity effects on shock attenuation, compaction and energy dissipation.

\textbf{Architected lattice.} A 0.3175 cm thick tantalum flier impacts a two‑dimensional architected aluminum lattice (Fig.~\ref{fig:setup:b}). The architected lattice comprises a square grid of struts at 0.10 cm pitch, rotated by an \textit{angle} in [0$^\circ$,45$^\circ$]. The strut thickness is set by the nominal \textit{porosity} (10–90\%). A copper backer of the same thickness lies downstream. The \textit{shock speed} is varied between 0.1 and 0.4 cm/$\mu$s. The architected lattice’s periodic structure and tunable rotation angle enable controlled manipulation of shock propagation, reflection and focusing.

For both configurations, we solve the compressible Euler equations with appropriate equations of state using LLNL’s MARBL hydrocode\cite{anderson2018high,dobrev2011curvilinear}. We record, at fixed intervals, seven fields—density $\rho$, pressure $p$, material indicator $m$, total energy $E$, temperature $T$, and two in-plane velocity components—on a $60\times60$ grid. These correspond to $u_r$ and $u_z$ for the porous case, and to $u_x$ and $u_y$ for the architected lattice case. These data constitute the training, validation and test sets for our deep‑learning model; additional details on the governing equations and numerical implementation appear in \ref{sec:gov_eqs} and \ref{sec:num_methods}.

\section{Deep Learning Model}
\label{sec:level5}

We employ a hybrid architecture combining convolutional neural networks (CNNs)\cite{lecun1998gradient} and long short‑term memory (LSTM) networks\cite{hochreiter1997long} to model the spatio‑temporal evolution of seven fields—pressure, temperature, energy, density, material indicator, and two velocity components—in 2D shock problems. CNNs extract local spatial patterns such as wavefronts and pore collapses, whereas LSTMs capture temporal dependencies, enabling the model to integrate features over multiple time steps.

\begin{figure}[!ht]
  \centering
  \includegraphics[width=0.7\textwidth]{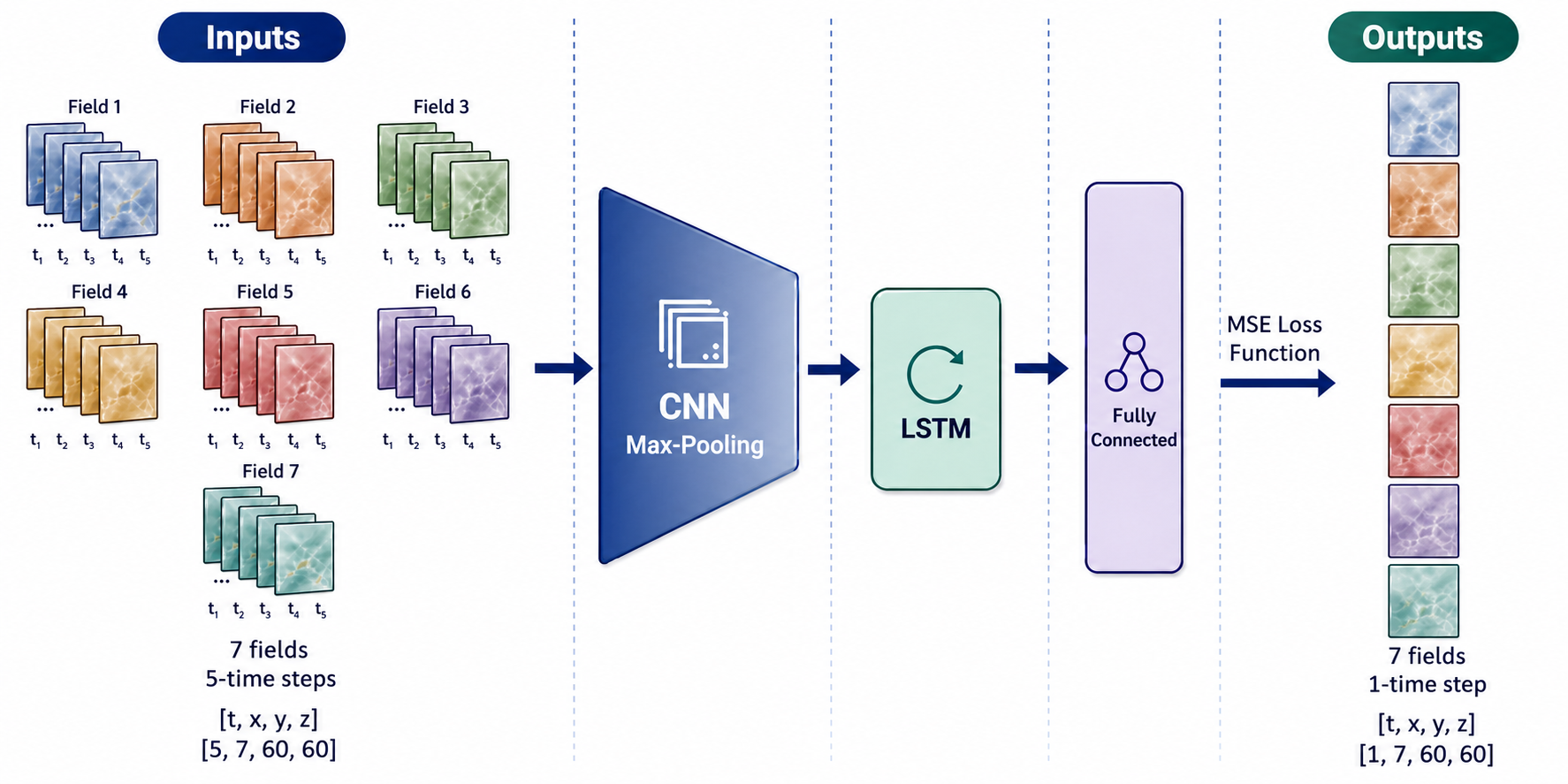}
  \caption{MSTM schematic architecture. The model combines CNN and LSTM blocks to recursively predict the seven‑field evolution. The CNN layers extract spatial features at each time step, and the LSTM layers capture their temporal evolution.}
  \label{fig:label02}
\end{figure}

\begin{figure}[!ht]
 \centering
 \includegraphics[width=0.52\textwidth]{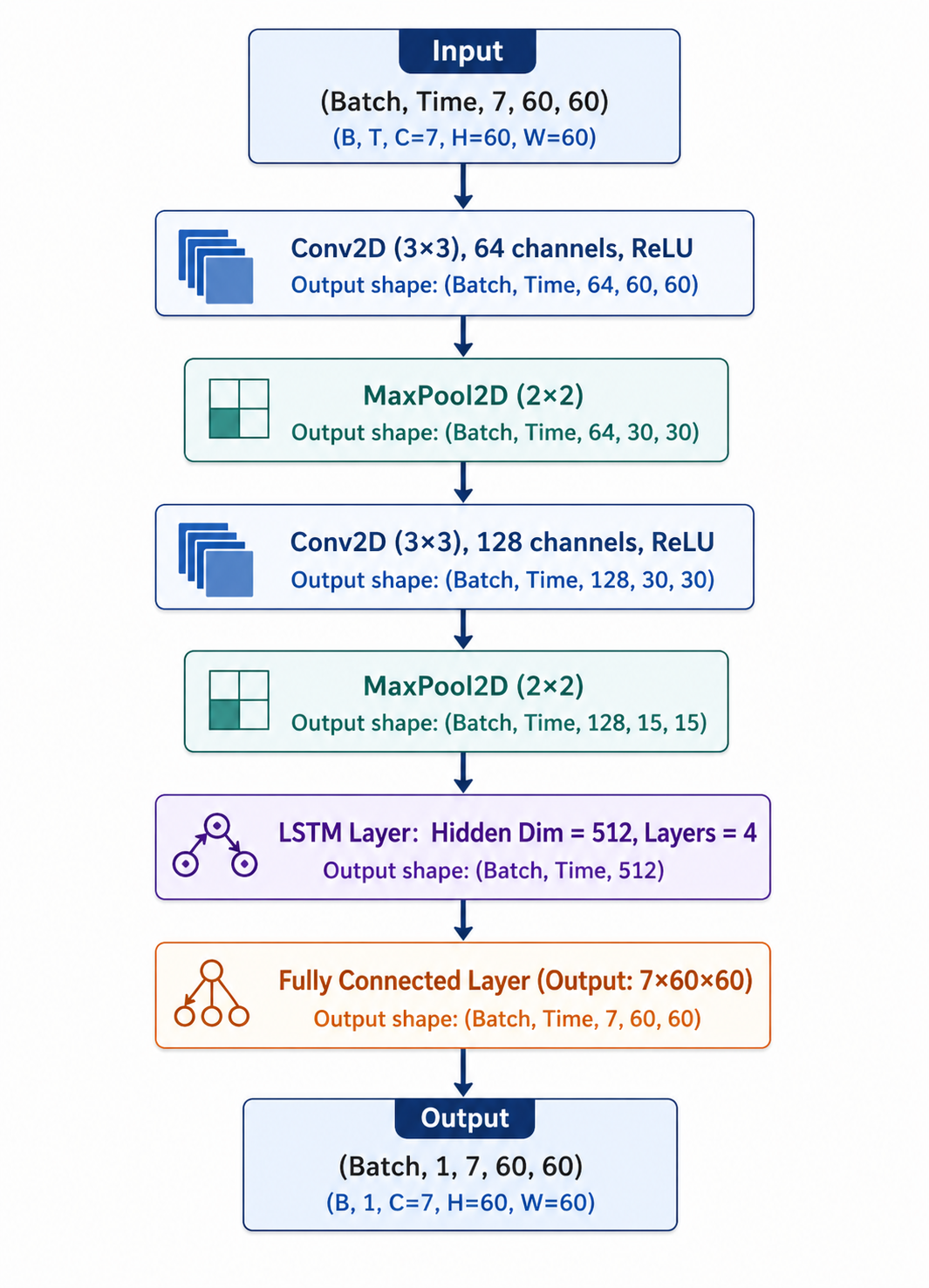}
 \caption{Formal description of the MSTM architecture. The model takes as input a batch of sequences with shape ($\text{batch\_size}$, T, 7, 60, 60), processes each slice through two convolutional layers (3$\times$3 kernels and ReLU activations) and max pooling, feeds the resulting feature vectors into a four‑layer LSTM, and uses a fully connected layer to project the final hidden state to the next‑step field tensor $(1,7,60,60)$.}
 \label{fig:label03}
\end{figure}

\begin{figure}[!ht] \centering \includegraphics[width=0.80\textwidth]{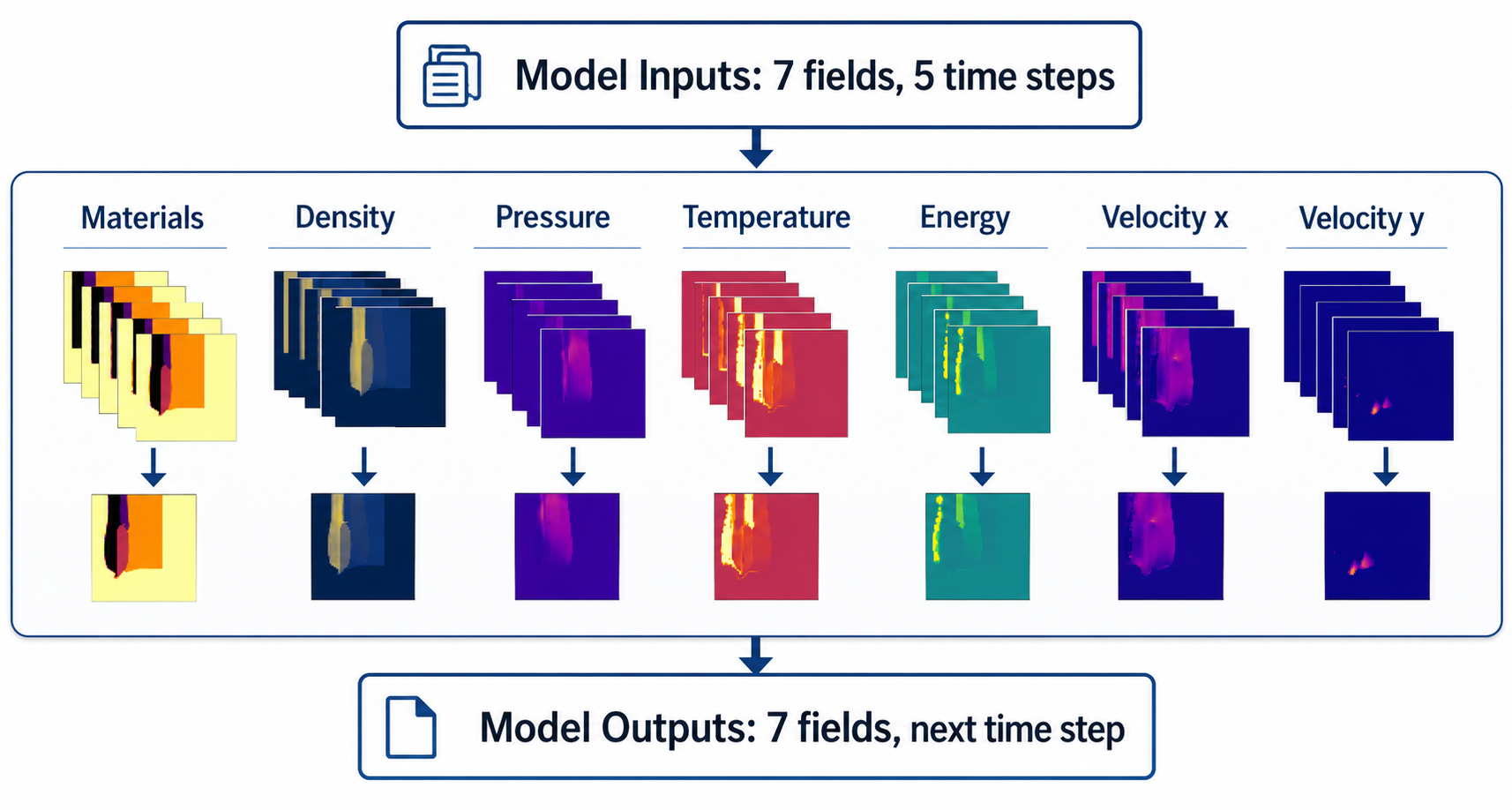} \caption{Single‐step field prediction of the porous material MSTM model. The upper panel shows the model inputs: five consecutive time steps for each of the seven physical fields on a \(60\times60\) grid (shape \([5,7,60,60]\)). Each field is visualized as a stack of five two‐dimensional snapshots; the arrows indicate how these temporal sequences feed into the model. The lower panel displays the model outputs: one predicted frame for each of the seven fields on the same \(60\times60\) grid at the next time step (shape \([1,7,60,60]\)).} \label{fig:label04} \end{figure}

\textbf{Architecture.} Figure \ref{fig:label02} presents a schematic of MSTM. Each training sample consists of five consecutive time frames, each with seven fields on a $60\times60$ grid. The feature extractor contains two convolutional layers: a $3\times3$ Conv2d layer with 64 output channels (\text{padding}=1) followed by $2\times2$ max pooling, and a second $3\times3$ Conv2d with 128 channels and another $2\times2$ max pool. These layers reduce the spatial resolution to $15\times15$ while preserving the receptive field, yielding spatial features of size $128\times15\times15$ at each time step. The flattened features are processed sequentially by a four‑layer LSTM with 512 hidden units; its final hidden state is passed through a fully connected layer that outputs a tensor of shape $(1,7,60,60)$ representing the next time step’s fields. Figure \ref{fig:label03} provides a formal diagram of this pipeline, while Fig. \ref{fig:label04} visualizes a single‑step prediction process for the porous case.

\textbf{Training procedure.} We train the model on data generated by the MARBL high-fidelity hydrocode. Each sample comprises five input frames and one target frame (the sixth). We normalize each field to the range $[0,1]$ using min-max scaling based on training-set extrema and apply the same parameters to validation and test data. We use mean squared error (MSE) as the training loss in the present work because it places greater weight on large local deviations than mean absolute error (MAE), which is useful in the presence of sharp shock-induced gradients. In our experiments, we found that MAE and cross-entropy losses converge more slowly and yield lower accuracy. In this initial formulation, conservation properties are assessed post hoc using metrics such as relative conservation of mass rather than being enforced directly in the training objective. Physics-informed loss formulations remain an important direction for future work. Optimization employs the Adam algorithm with learning rate $5\times10^{-4}$ and \texttt{batch\_size} 256. We train for 1{,}000 epochs on LLNL’s Lassen supercomputer using PyTorch’s \texttt{DataParallel} on four NVIDIA Volta GPUs. Training takes roughly 7 hours for the porous dataset (679 training, 85 validation, and 91 test sequences) and 24 hours for the architected lattice dataset (2{,}387 training, 298 validation, and 325 test sequences). The train/validation/test split is approximately 80/10/10\%. Because total training time scales with sequence count, the larger architected lattice dataset requires longer training.

\textbf{Choice of a five‑frame window.} The input window spans five time frames. This span (1.0 $\mu$s for the porous simulations) was chosen to match the dominant shock transit time across mesoscale features. For the porous configuration, a shock traveling at 0.23 cm/$\mu$s crosses approximately 0.2–0.25 mm during this interval, comparable to ligament thickness in the porous aluminum. For architected lattice runs, the window spans 0.05–0.5 $\mu$s (covering 50 frames) depending on shock speed; during this interval the shock front traverses 0.05–0.2 mm, comparable to the 0.10 cm unit‑cell pitch and typical strut thickness. During inference, this five-frame context is advanced autoregressively to generate the entire simulation sequence (60 frames for porous and 50 for architected lattice cases). Shorter windows may truncate relevant temporal information before the shock front interacts fully with microstructural features, whereas longer windows would increase memory requirements and training complexity. A systematic ablation over temporal window length and recurrent architecture, including LSTM depth and hidden dimension, is left for future work. This balanced choice thus captures the onset and progression of pore collapse, wave interactions and other nonlinearities while keeping the autoregressive rollout stable and computationally tractable.

\begin{figure}[!ht]
    \centering
    \includegraphics[width=0.8\textwidth]{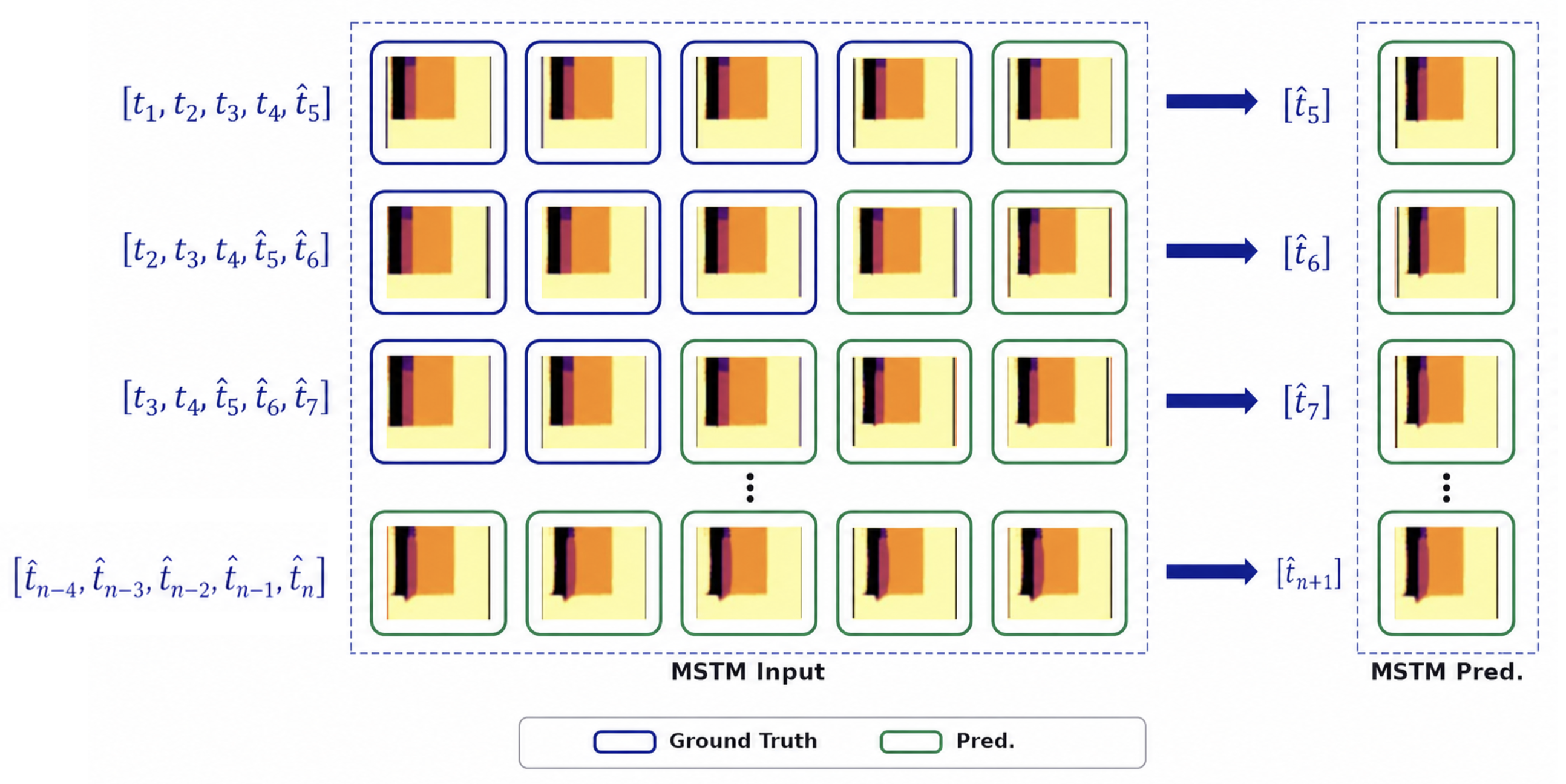}
    \caption{Autoregressive inference process of the MSTM model for the material indicator field in the porous case. In the first step, the network receives five consecutive ground‑truth frames $z_{t_0}\ldots z_{t_4}$ (blue outlines) to predict frame $\hat{z}_{t_5}$ (green). For the next prediction, the input window advances by discarding $z_{t_0}$ and appending the most recent prediction, yielding $\{z_{t_1},\ldots,z_{t_4},\hat{z}_{t_5}\}$ to produce $\hat{z}_{t_6}$. This sliding‑window process continues, each new prediction replaces the oldest entry in the input sequence, until the final target frame $\hat{z}_{t_{n+1}}$ is generated.}
    \label{fig:label05}
\end{figure}

\textbf{Teacher forcing and autoregressive rollout.} During training, we employ teacher forcing: the model receives five consecutive ground truth frames $\{z_{t_0},\dots,z_{t_4}\}$ and predicts $\hat{z}_{t_5}$, without feeding its own outputs back. During inference, we seed the network with the first five true frames and then enter an autoregressive regime: each new prediction replaces the oldest frame in the input sequence until the end of the sequence (see Fig.~\ref{fig:label05}). This scheme enables the model to generate full spatio-temporal rollouts for up to 60 time steps in the porous case and 50 in the architected lattice case. This training-inference mismatch can introduce exposure bias, whereby small errors accumulate when the model is conditioned on its own predictions during rollout. In the present results, this effect is more visible in the architected lattice cases, where thin struts and sharp interfaces are more sensitive to small phase shifts. Scheduled sampling, in which the model is gradually exposed during training to its own predictions rather than only to ground truth histories, is a possible extension for improving long-rollout stability.

\textbf{Implementation.} The MSTM model is implemented in PyTorch and parallelized across GPUs to accelerate training. The layer-by-layer specification, hyperparameters and implementation details are summarized in Table~\ref{table:architecture.} and ~\ref{app:architecture}.

\section{Results}
\label{sec:results}
\subsection{Qualitative Performance}

\begin{figure*}[h!]
  \centering
  \begin{tabular}{cc}
    \subcaptionbox{Material indicator\label{fig:materials}}{%
      \includegraphics[width=0.48\textwidth]{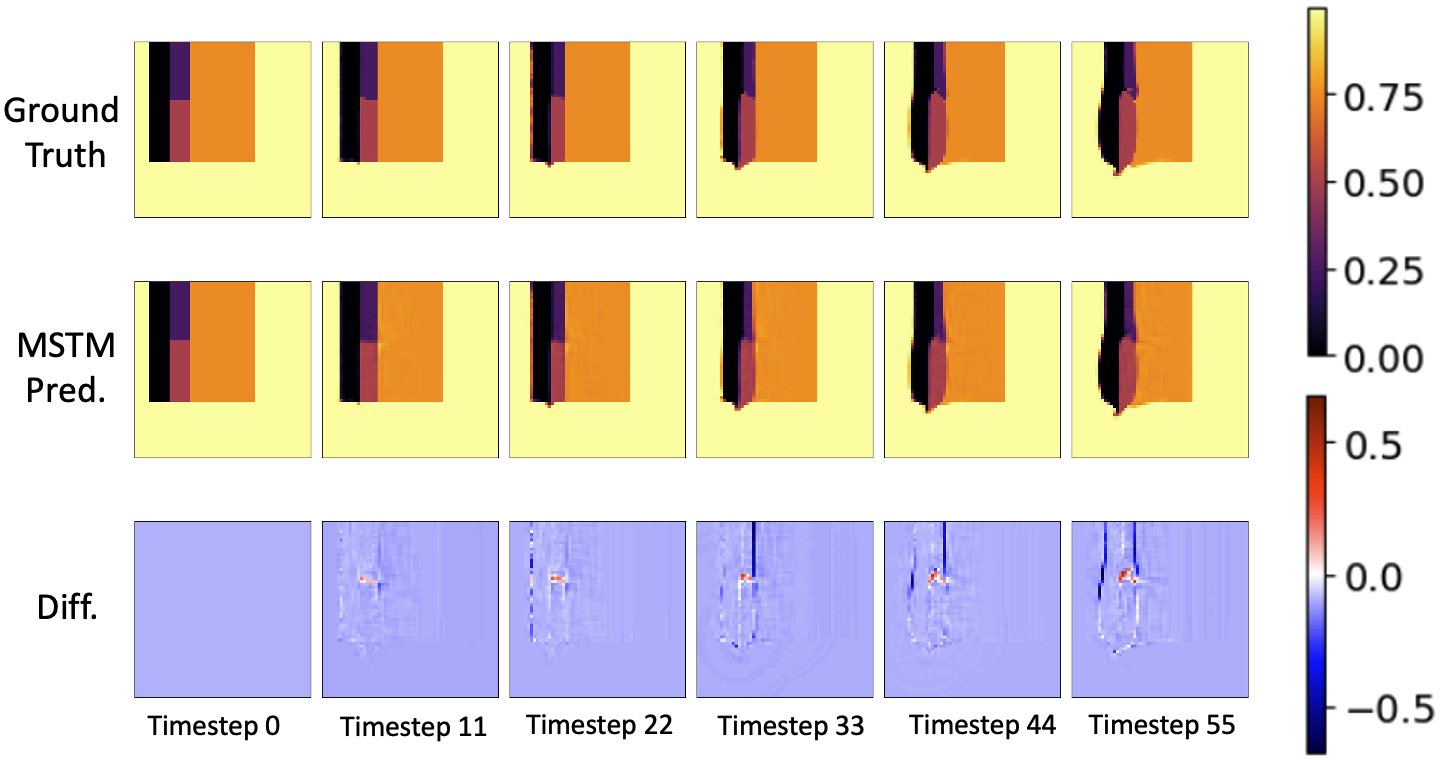}} &
    \subcaptionbox{Density\label{fig:density}}{%
      \includegraphics[width=0.48\textwidth]{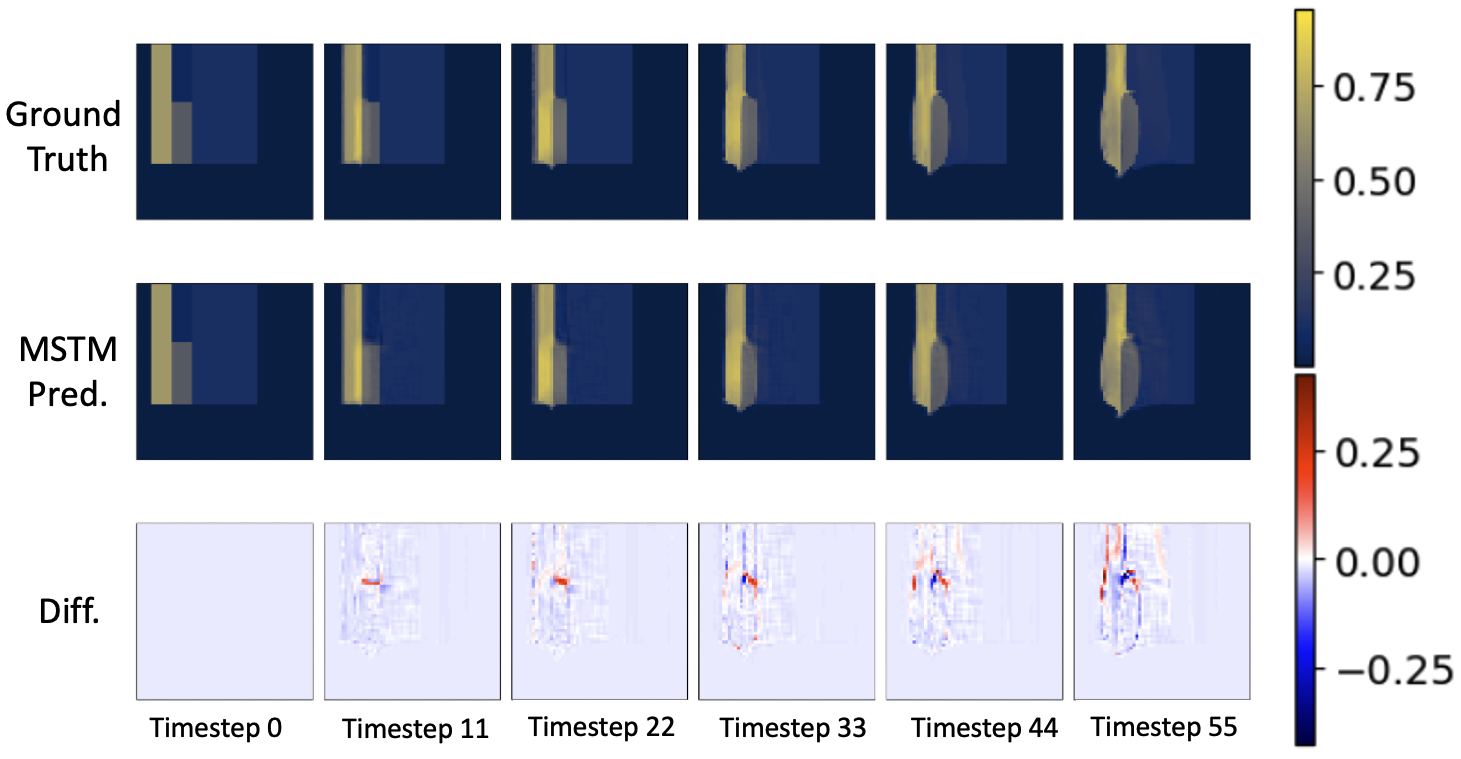}} \\[1ex]
    \subcaptionbox{Velocity in $z$\label{fig:velocity_x}}{%
      \includegraphics[width=0.48\textwidth]{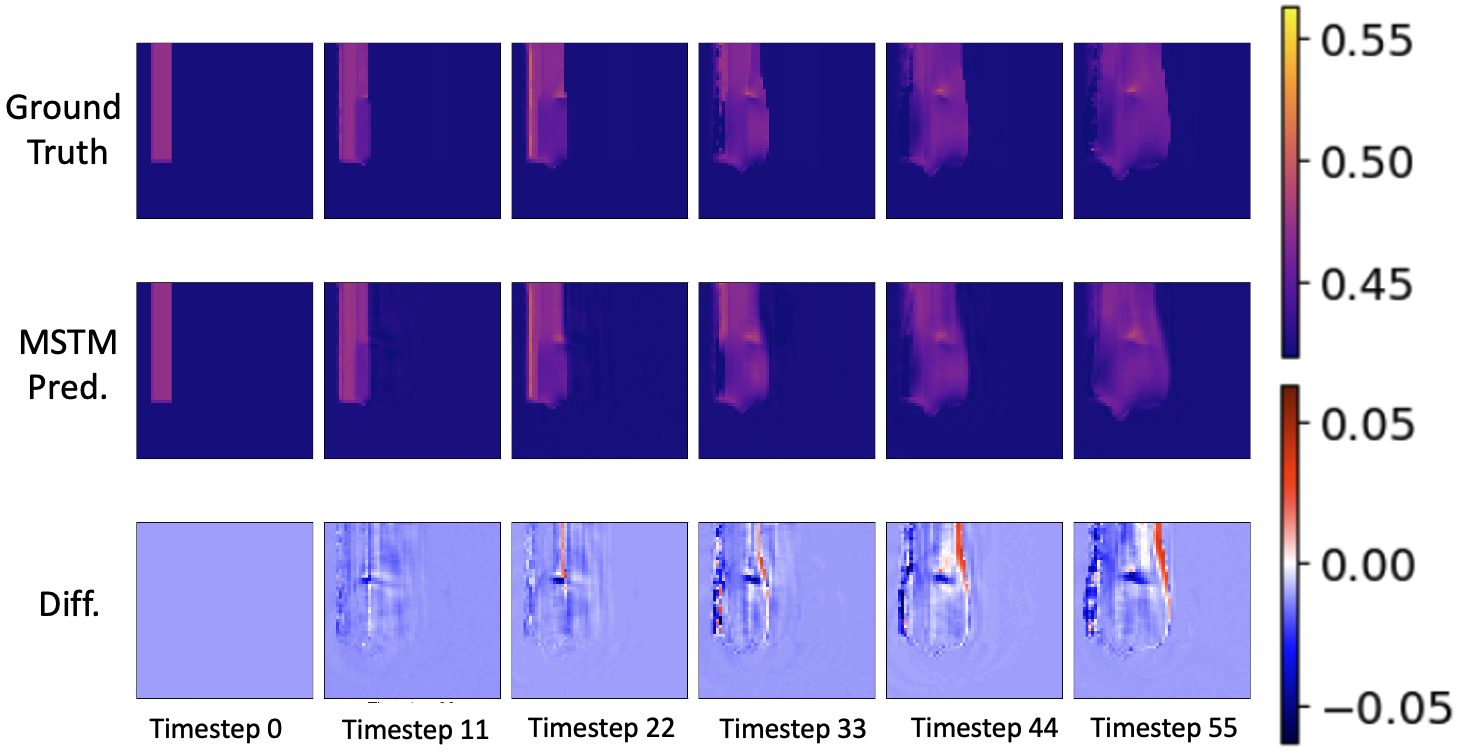}} &
    \subcaptionbox{Velocity in $r$\label{fig:velocity_y}}{%
      \includegraphics[width=0.48\textwidth]{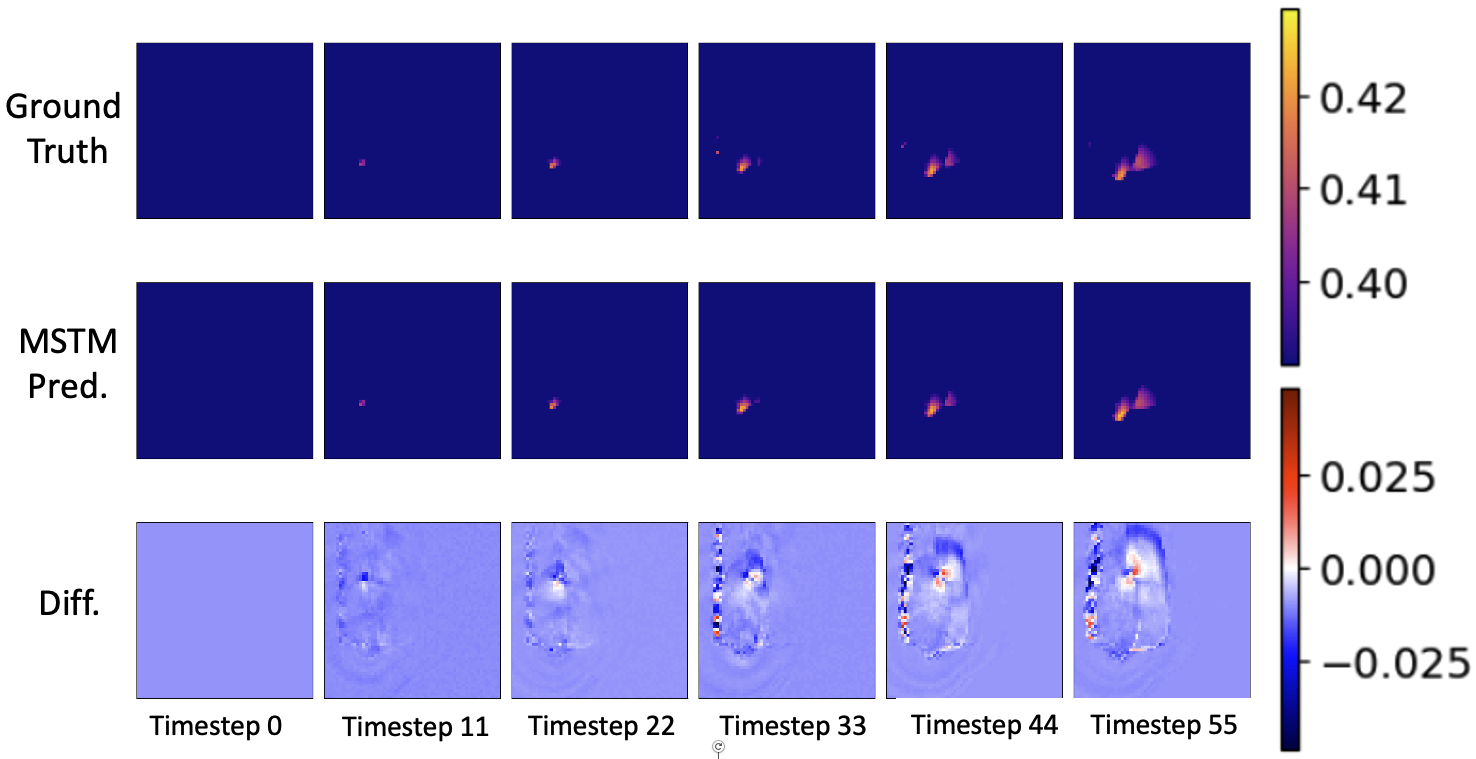}} \\[1ex]
    \subcaptionbox{Energy\label{fig:energy}}{%
      \includegraphics[width=0.48\textwidth]{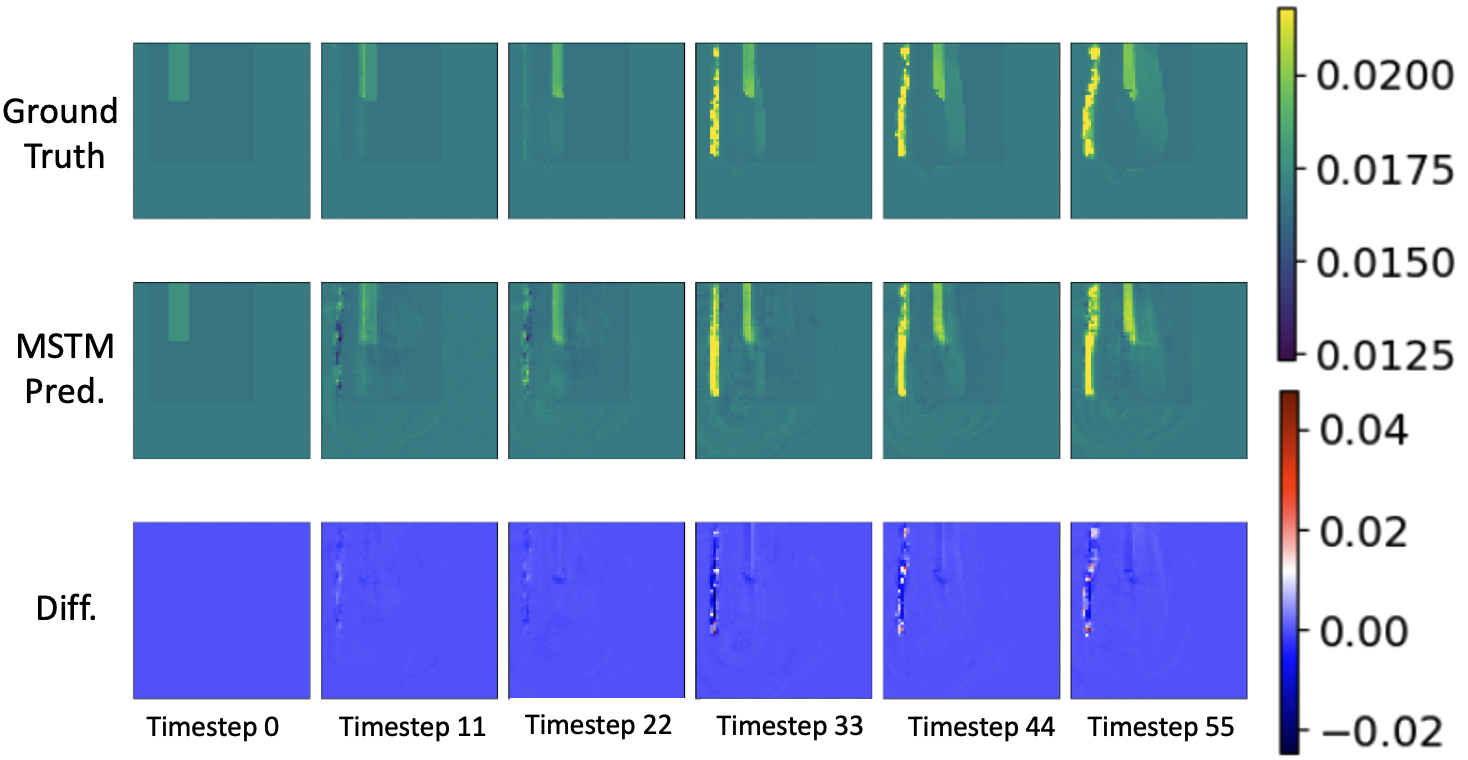}} &
    \subcaptionbox{Pressure\label{fig:pressure}}{%
      \includegraphics[width=0.48\textwidth]{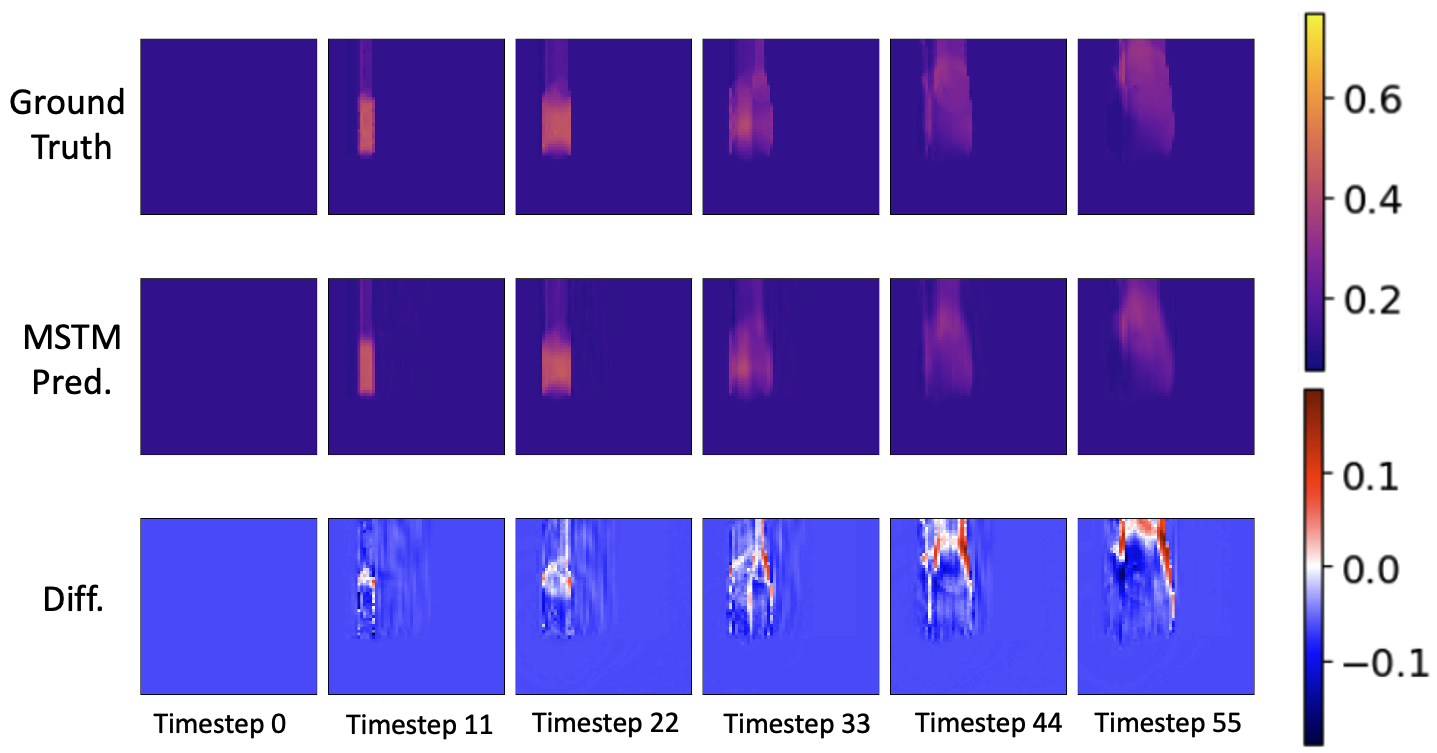}} \\[1ex]
    \multicolumn{2}{c}{%
      \subcaptionbox{Temperature\label{fig:temperature}}{%
        \includegraphics[width=0.48\textwidth]{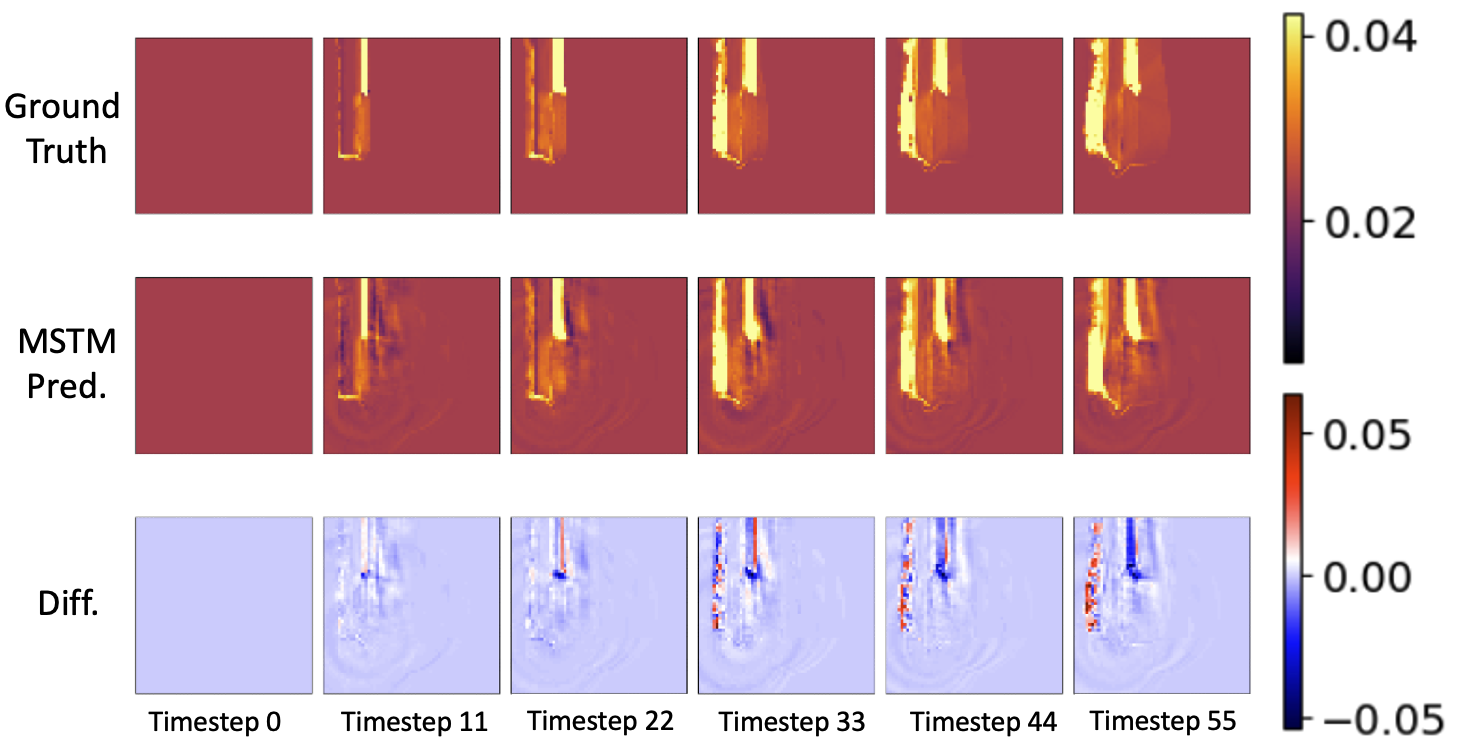}}}
  \end{tabular}
  
  \caption{Qualitative comparison of MSTM predictions against high‑fidelity porous material simulations for all seven physical fields. Each subfigure displays six evenly spaced time steps (0, 11, 22, 33, 44, 55). Within each subfigure, the top row shows the ground‑truth field, the middle row shows the corresponding MSTM prediction, and the bottom row shows the difference (prediction minus ground truth). Panels (\ref{fig:materials}) and (\ref{fig:density}) present the material indicator and density, panels (\ref{fig:velocity_x}) and (\ref{fig:velocity_y}) show velocity in \(z\) and \(r\), panel (\ref{fig:energy}) shows energy, panel (\ref{fig:pressure}) shows pressure, and panel (\ref{fig:temperature}), beneath, shows temperature. Color bars at the right of each block indicate the data range for ground truth and prediction, while the difference row uses a diverging scale to highlight local errors.}
  \label{fig:all-fields}
\end{figure*}

\begin{figure*}[h!]
  \centering
  \begin{tabular}{cc}
    \subcaptionbox{Material indicator\label{fig:materials2}}{%
      \includegraphics[width=0.48\textwidth]{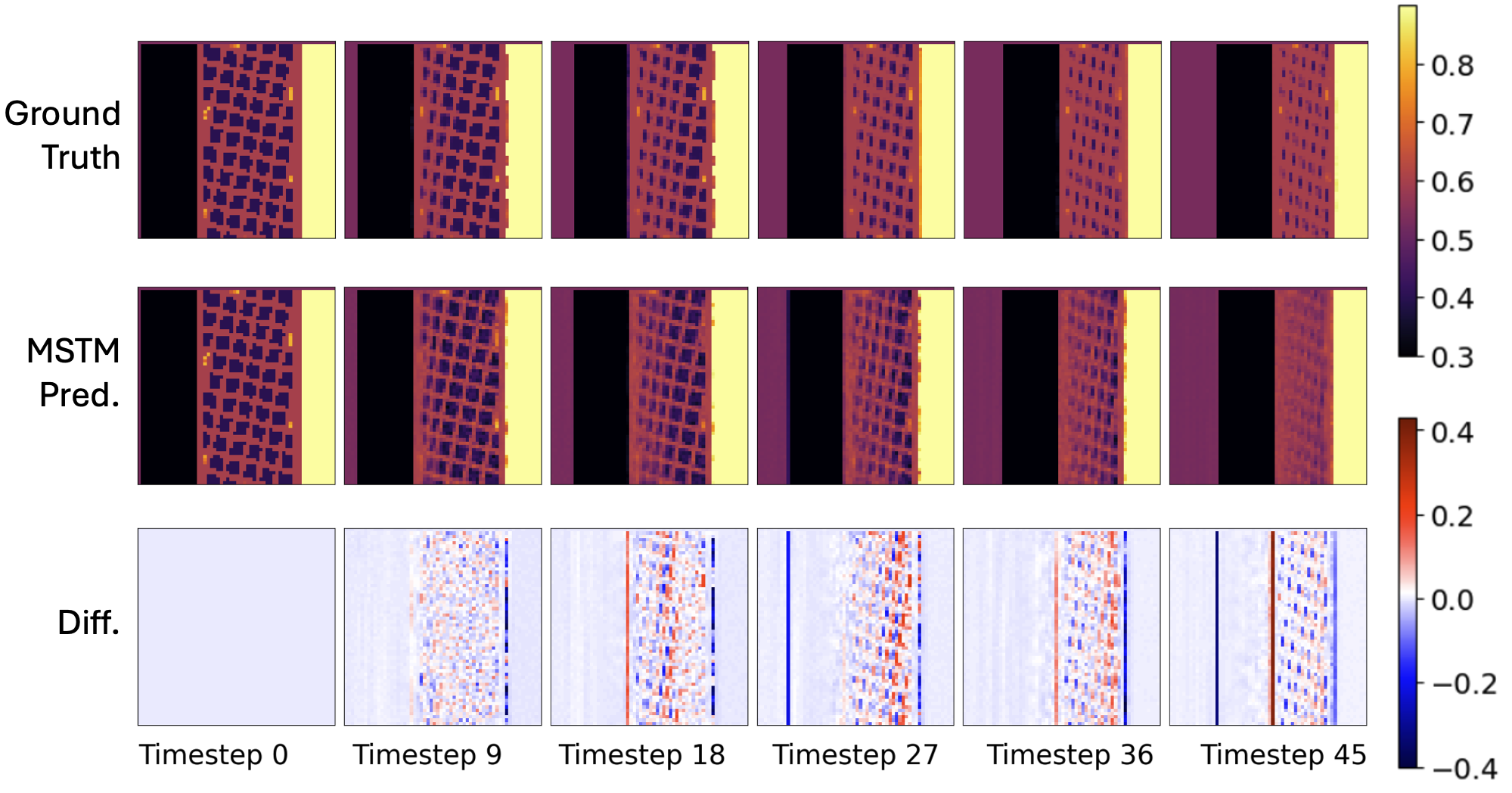}} &
    \subcaptionbox{Density\label{fig:density2}}{%
      \includegraphics[width=0.48\textwidth]{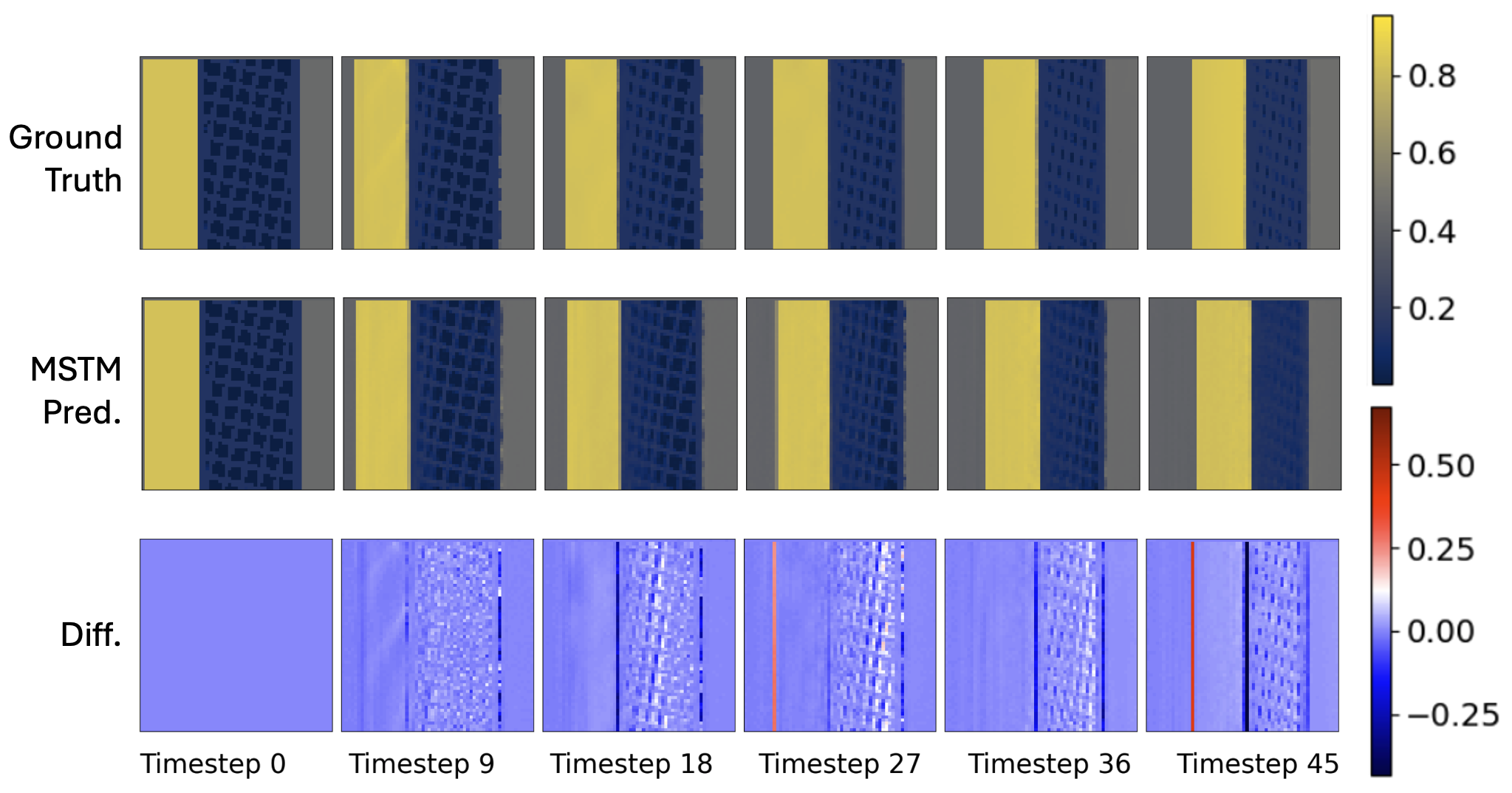}} \\[1ex]
    \subcaptionbox{Velocity in $x$\label{fig:velocity_x2}}{%
      \includegraphics[width=0.48\textwidth]{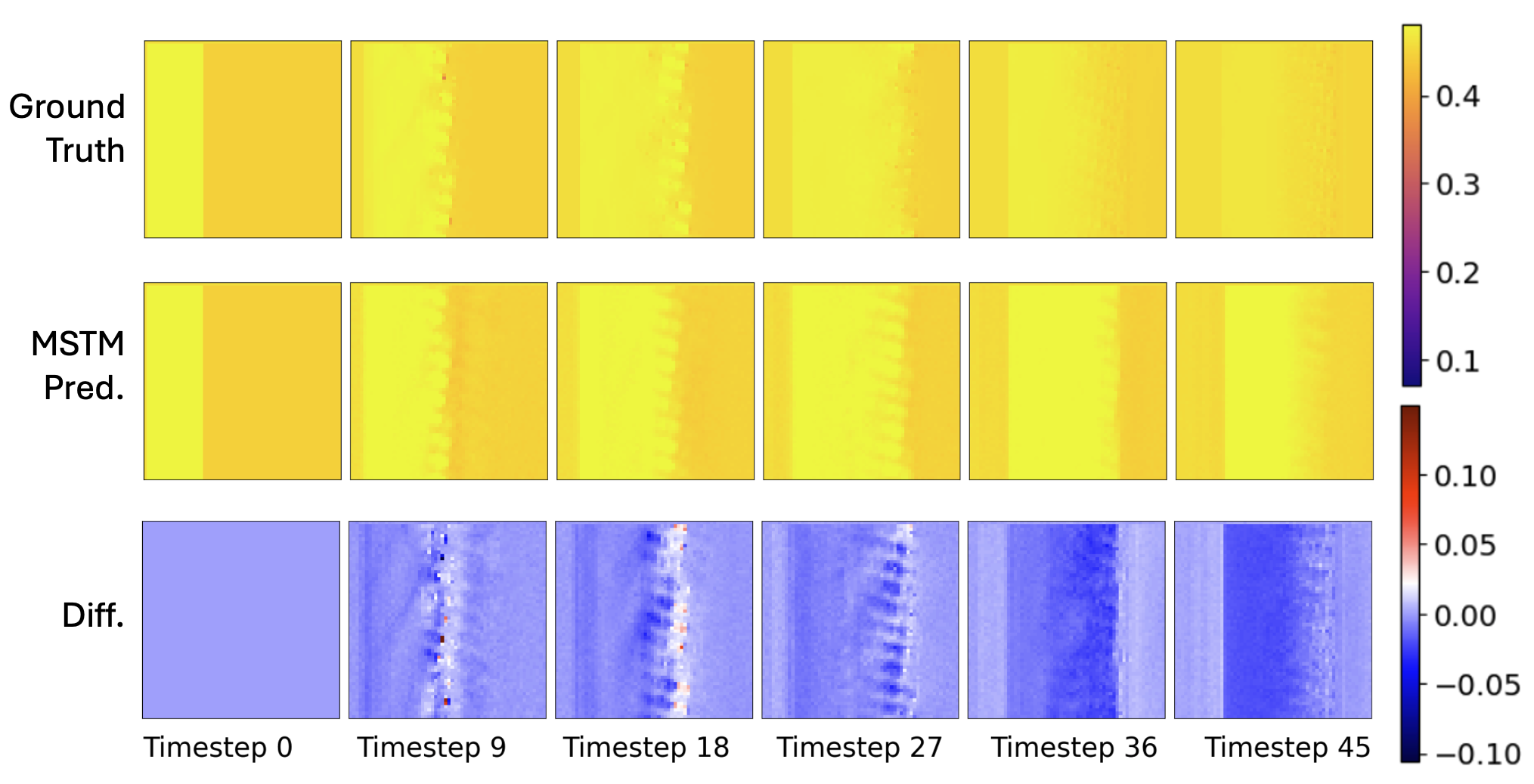}} &
    \subcaptionbox{Velocity in $y$\label{fig:velocity_y2}}{%
      \includegraphics[width=0.48\textwidth]{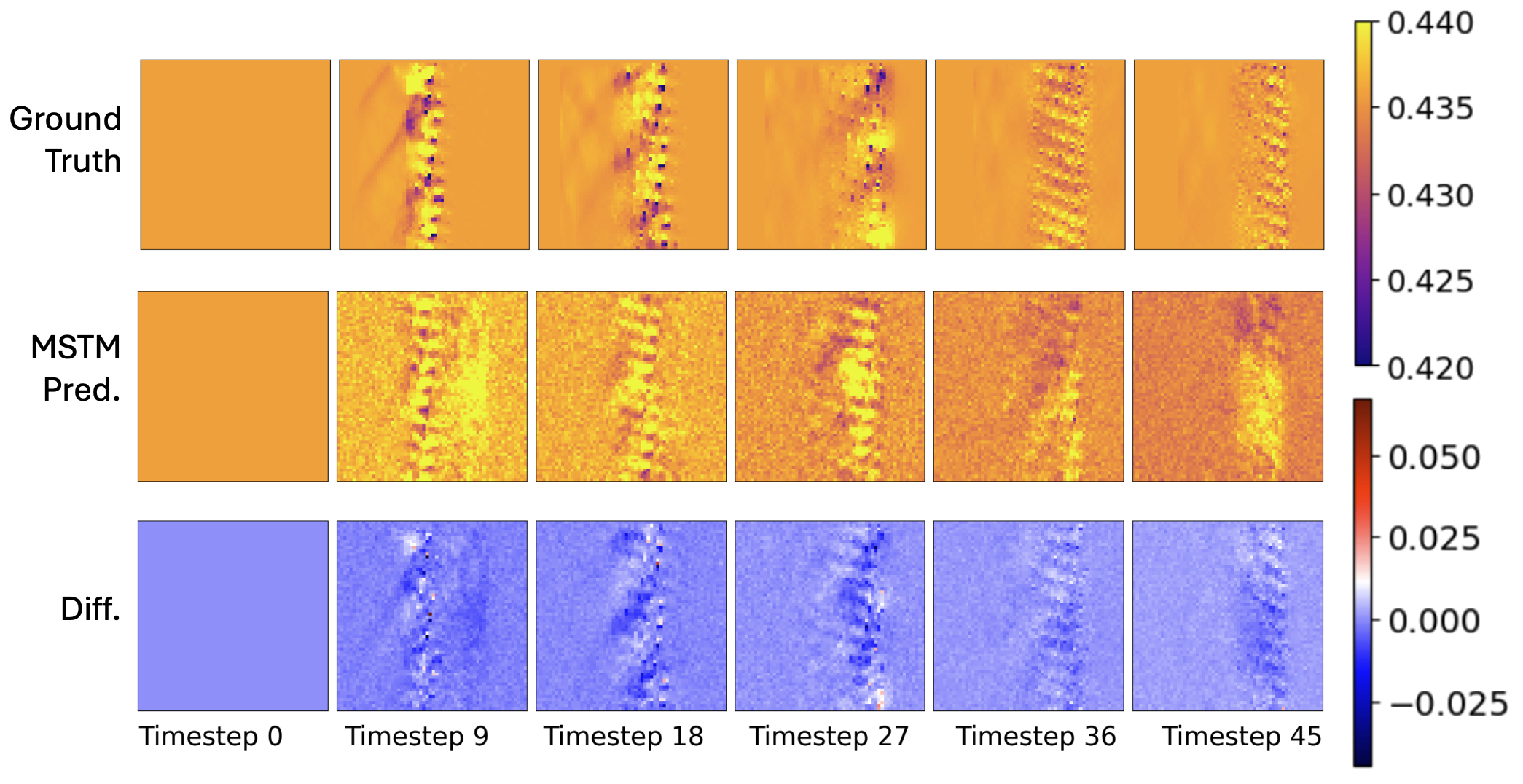}} \\[1ex]
    \subcaptionbox{Energy\label{fig:energy2}}{%
      \includegraphics[width=0.48\textwidth]{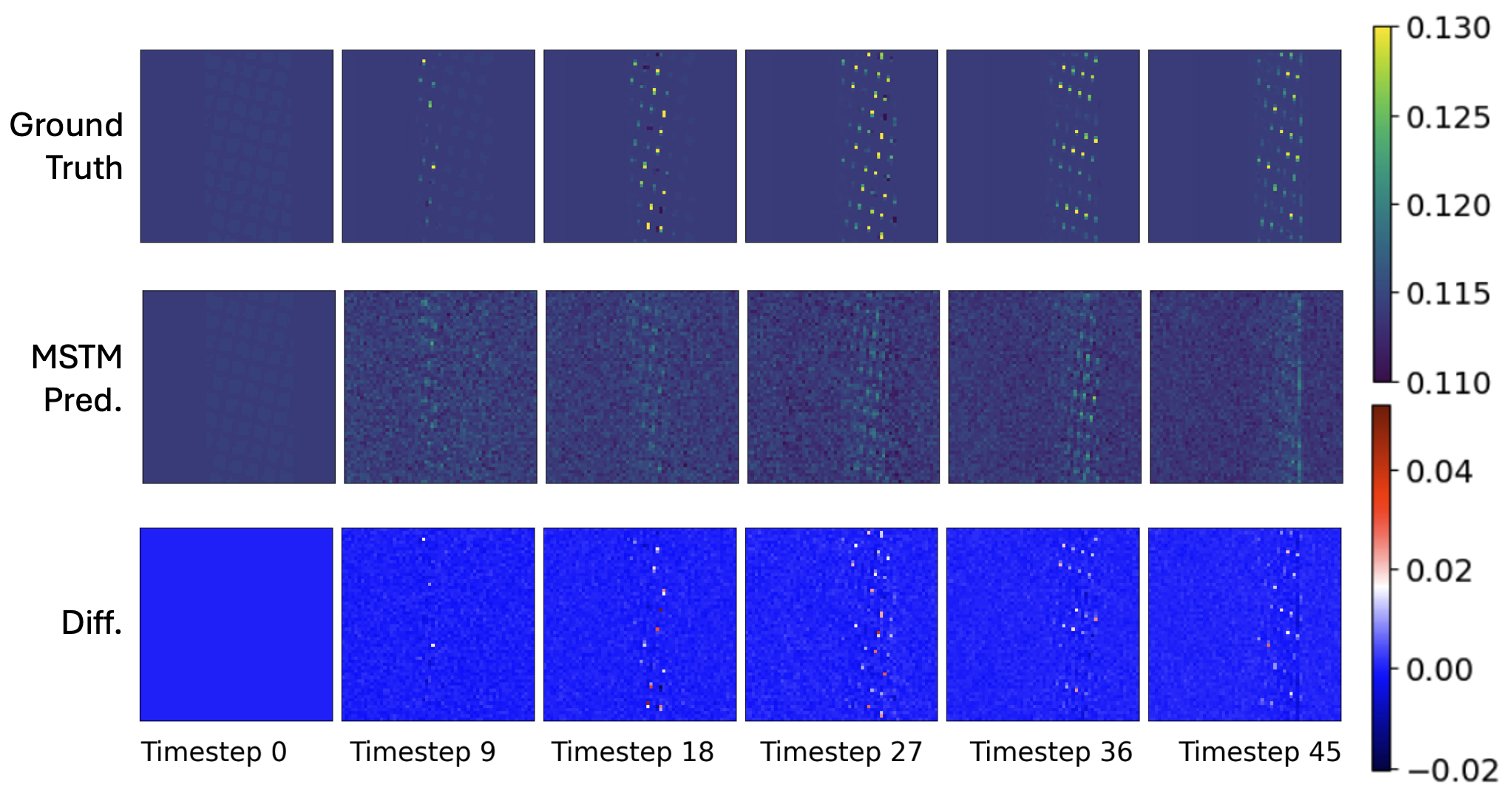}} &
    \subcaptionbox{Pressure\label{fig:pressure2}}{%
      \includegraphics[width=0.48\textwidth]{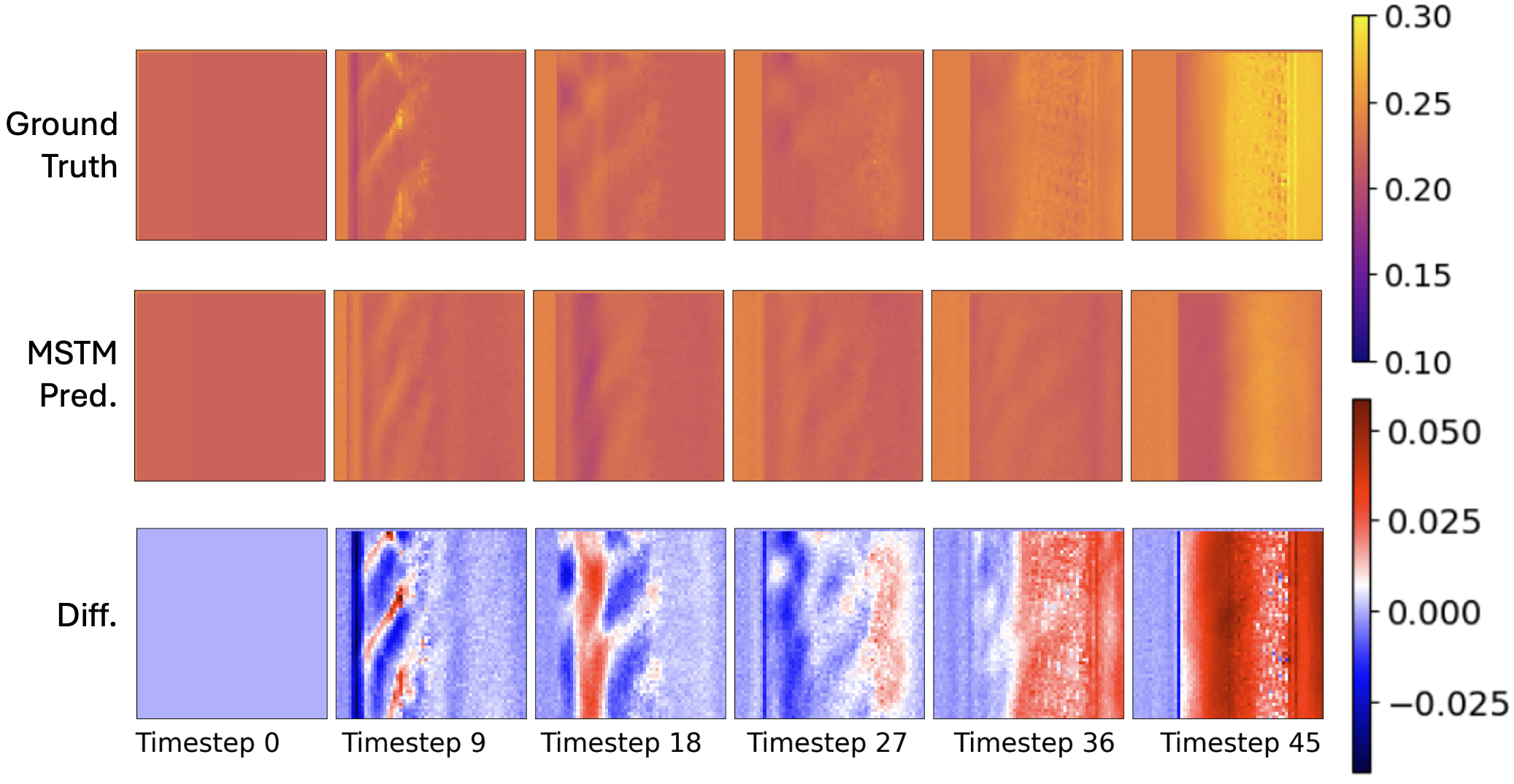}} \\[1ex]
    \multicolumn{2}{c}{%
      \subcaptionbox{Temperature\label{fig:temperature2}}{%
        \includegraphics[width=0.48\textwidth]{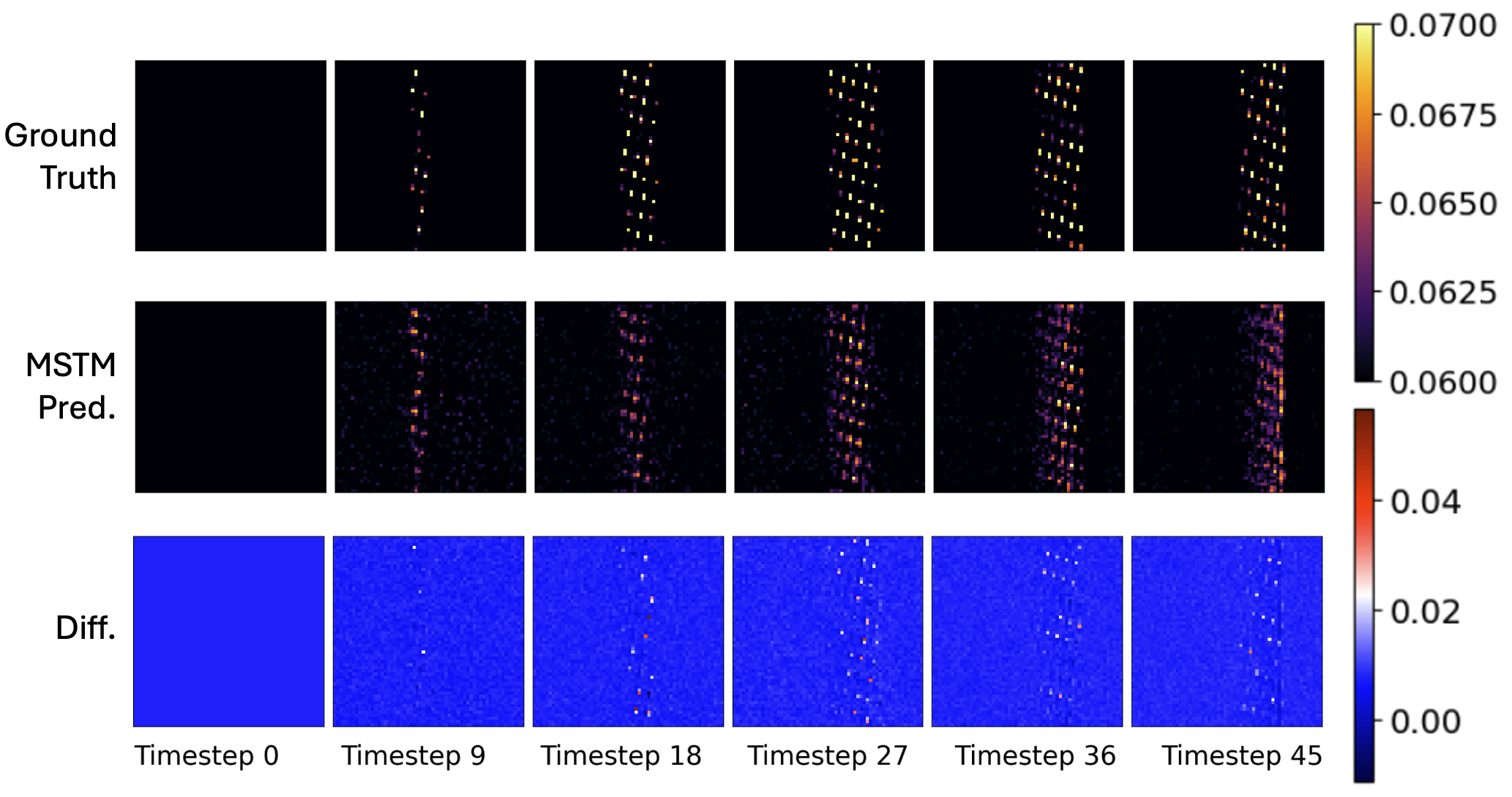}}}
  \end{tabular}
  \caption{Qualitative comparison of MSTM predictions against high-fidelity architected lattice simulations for all seven physical fields. Each subfigure shows six evenly spaced time steps (0, 9, 18, 27, 36, 45). Within each panel, the top row is the ground truth field, the middle row is the corresponding MSTM prediction, and the bottom row is the difference (prediction minus ground truth). Panels (\ref{fig:materials2}) and (\ref{fig:density2}) display the material indicator and density, panels (\ref{fig:velocity_x2}) and (\ref{fig:velocity_y2}) show velocity in \(x\) and \(y\), panels (\ref{fig:energy2}) and (\ref{fig:pressure2}) show energy and pressure in a two-column arrangement, and panel (\ref{fig:temperature2}), beneath, shows temperature. Color bars to the right of each block indicate absolute field values, while the difference rows use a diverging color scale to highlight local errors.}
  \label{fig:all-fields-lattice}
\end{figure*}

Figures \ref{fig:all-fields} and \ref{fig:all-fields-lattice} provide a qualitative assessment of MSTM in predicting the multi-field response to the shock over time. The figures feature a single test sequence, randomly selected, which was not seen by the model during training and validation. The first row represents the ground truth, which is the actual field from the simulation. Note that each row shows six snapshots drawn at evenly spaced intervals throughout the simulated sequence. The second row shows the MSTM predictions for the same time steps. Visually, the predictions closely follow the ground truth, capturing the overall field evolution in both porous and architected lattice configurations. The third row presents the difference between the ground truth and MSTM predictions. Most of the differences are small, but some challenging areas appear around shock interfaces, which are regions of rapid change. Overall, MSTM predictions remain in close agreement with the simulated field evolution throughout the rollout.

\subsection{Performance Metrics}

\begin{figure*}[!ht]
  \centering
  \begin{subfigure}[t]{0.35\textwidth}
    \centering
    \includegraphics[width=\linewidth]{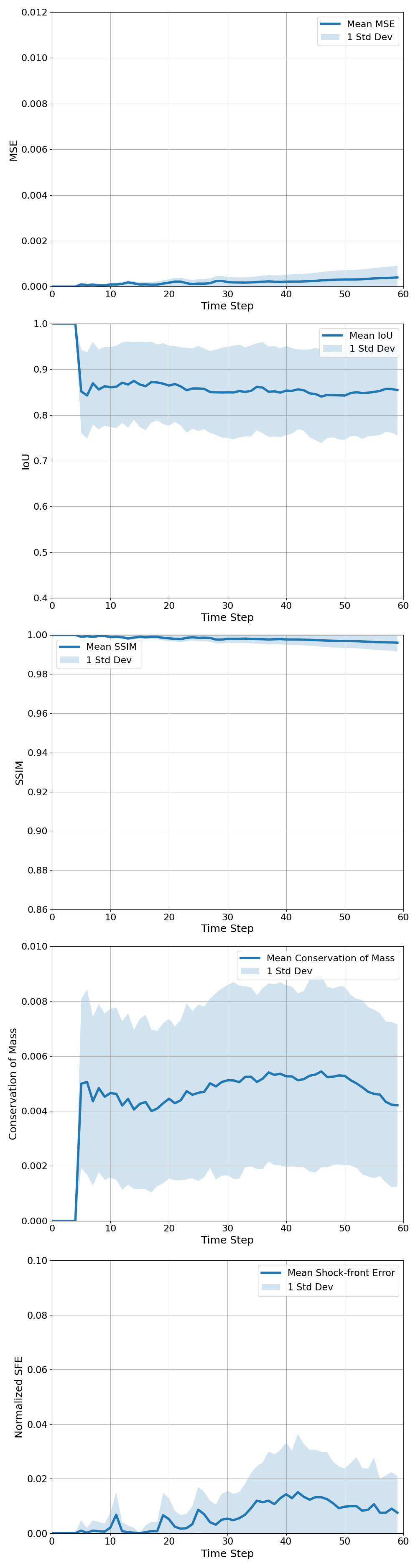}
    \caption{Performance over 91 porous test simulations.}
    \label{fig:metrics:porous}
  \end{subfigure}
  \hspace{0.02\textwidth}%
  \begin{subfigure}[t]{0.35\textwidth}
    \centering
    \includegraphics[width=\linewidth]{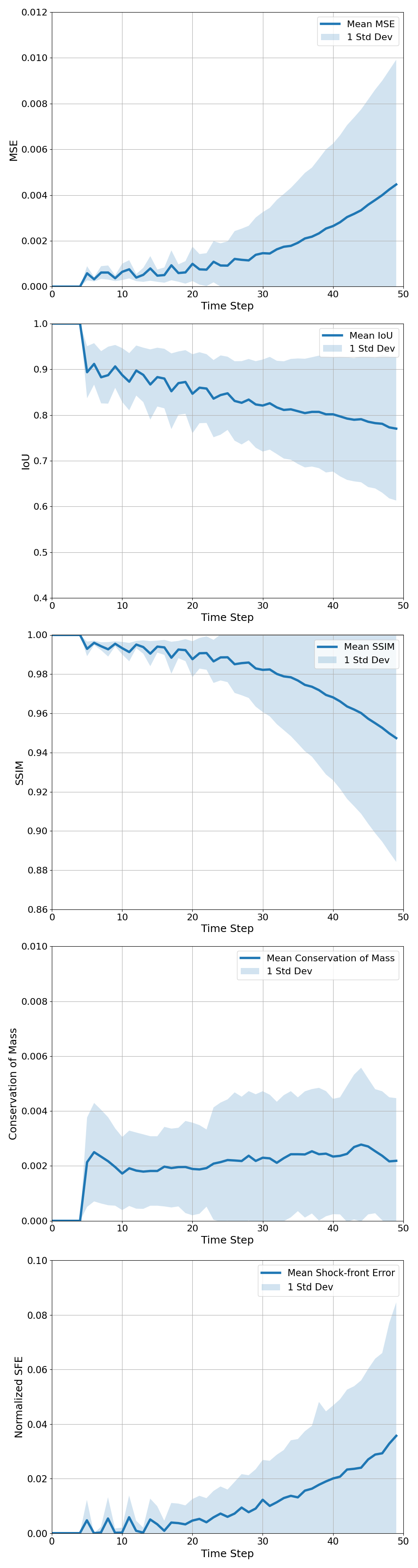}
    \caption{Performance over 325 architected lattice test simulations.}
    \label{fig:metrics:lattice}
  \end{subfigure}
  \caption{Temporal evolution of performance metrics for MSTM on (a) porous and (b) architected lattice test simulations. Each panel contains five stacked metrics: mean squared error (MSE), intersection-over-union (IoU), structural similarity index (SSIM), relative mass error (CM), and pressure-based normalized shock-front location error ($\mathrm{SFE}_{\mathrm{norm}}$). Solid curves denote means across test runs and shaded regions denote one standard deviation; the plots label these bands as 1 Std Dev. The $\mathrm{SFE}_{\mathrm{norm}}$ row reports front-location error normalized by the 60-cell axial domain length. The porous panel spans 60 time steps and the architected lattice panel spans 50 time steps. Identical axis ranges for corresponding metrics facilitate direct comparison.}
  \label{fig:metrics:combined}
\end{figure*}

To quantify MSTM’s predictive accuracy, we evaluate five complementary metrics: mean squared error (MSE), intersection over union (IoU) on the material indicator, structural similarity index measure (SSIM), relative conservation of mass (CM), and a pressure-based normalized shock-front location error ($\mathrm{SFE}_{\mathrm{norm}}$). \ref{app:metrics} details the formulas and averaging conventions. MSE measures the average squared difference between predicted and true field values\cite{goodfellow2016deep}, IoU quantifies overlap between predicted and true material masks using a soft interval membership\cite{jaccard1912distribution}, SSIM assesses perceptual similarity in luminance, contrast and structure\cite{wang2004image}, CM computes the relative error in total mass to check conservation properties\cite{leveque2002finite}, and $\mathrm{SFE}_{\mathrm{norm}}$ measures the absolute difference between predicted and true pressure-front positions normalized by the 60-cell axial domain length. The reported metric values are computed on the normalized fields used by the network, with each field scaled to $[0,1]$ using training-set extrema. Figure~\ref{fig:metrics:combined} plots the time evolution of MSE, IoU, SSIM, CM, and $\mathrm{SFE}_{\mathrm{norm}}$ for 91 porous and 325 architected lattice test simulations (panels a and b, respectively), while Table~\ref{metrics_table} summarizes the corresponding aggregate performance metrics. MSE, RMSE, SSIM, IoU, and CM are computed on the normalized fields used by the network, and $\mathrm{SFE}_{\mathrm{norm}}$ is reported as the front-location error divided by the 60-cell axial domain length.

Table~\ref{metrics_table} reports that MSTM achieves an average MSE of $2\times10^{-4}$ (RMSE $\approx1.4\%$) on the porous dataset and $1\times10^{-3}$ (RMSE $\approx3.2\%$) on the architected lattice dataset. IoU remains close to 0.85 in porous cases and declines to 0.78 over the architected lattice rollouts, reflecting the greater geometric complexity and sensitivity of the architected lattice mask to phase shifts. SSIM stays above 0.998 for porous runs and above 0.98 for architected lattice runs, indicating preserved luminance, contrast, and structure. CM errors remain below $5\times10^{-3}$ in porous runs and below $3\times10^{-3}$ in architected lattice runs, showing that although the lattice case is more challenging overall, some global quantities are slightly better preserved. The $\mathrm{SFE}_{\mathrm{norm}}$ row reports mean normalized front-location errors of $0.006$ for porous runs and $0.010$ for architected lattice runs, expressed as fractions of the domain length. Collectively, these metrics show that MSTM reproduces overall field values, fine-scale structures, and the dominant pressure-front propagation behavior across the studied test cases, even as errors accumulate modestly during autoregressive rollout in the more demanding architected lattice configurations.

\begin{table}[ht]
    \centering
    \caption{Performance metrics for the porous and architected lattice MSTM models, averaged over all time steps and test simulations. Values are mean $\pm$ one standard deviation. Arrows denote whether higher ($\uparrow$) or lower ($\downarrow$) is better. The $\mathrm{SFE}_{\mathrm{norm}}$ is reported as a normalized fraction of the 60-cell axial domain length.}
    \begin{tabular}{lcc}
        \toprule
        \textbf{Metric} & \textbf{Porous Model} & \textbf{Lattice Model} \\
        \midrule
        $\mathrm{MSE}\,\downarrow$ & $(2 \pm 2)\times10^{-4}$ & $(1\pm 1)\times10^{-3}$\\
        $\mathrm{IoU}\,\uparrow$   & $0.84\pm0.02$                   & $0.85\pm0.07$\\
        $\mathrm{SSIM}\,\uparrow$  & $0.998\pm0.002$                & $0.98\pm0.02$\\
        $\mathrm{CM}\,\downarrow$  & $(4 \pm 2)\times10^{-3}$ & $(2.0 \pm 0.7)\times10^{-3}$\\
        $\mathrm{SFE}_{\mathrm{norm}}\,\downarrow$ & $0.006 \pm 0.013$ & $0.010 \pm 0.022$\\
        \bottomrule
    \end{tabular}
    \label{metrics_table}
\end{table}

\subsection{Single-field versus multi-field model}
\label{app:single_vs_multi}

We quantify the value of cross-field coupling by training seven single-field spatio-temporal models (1STMs), one per field, under the same encoder, sequence length, optimizer, normalization, and rollout settings as the MSTM. Each 1STM receives the same five-frame context and predicts only its target field; MSTM receives and predicts all seven jointly. Evaluation uses the identical autoregressive protocol and metrics defined in the main text.

\begin{figure*}[h!]
  \centering
  \begin{tabular}{cc}
    \subcaptionbox{Material indicator\label{fig:app_materials}}{%
      \includegraphics[width=0.40\textwidth]{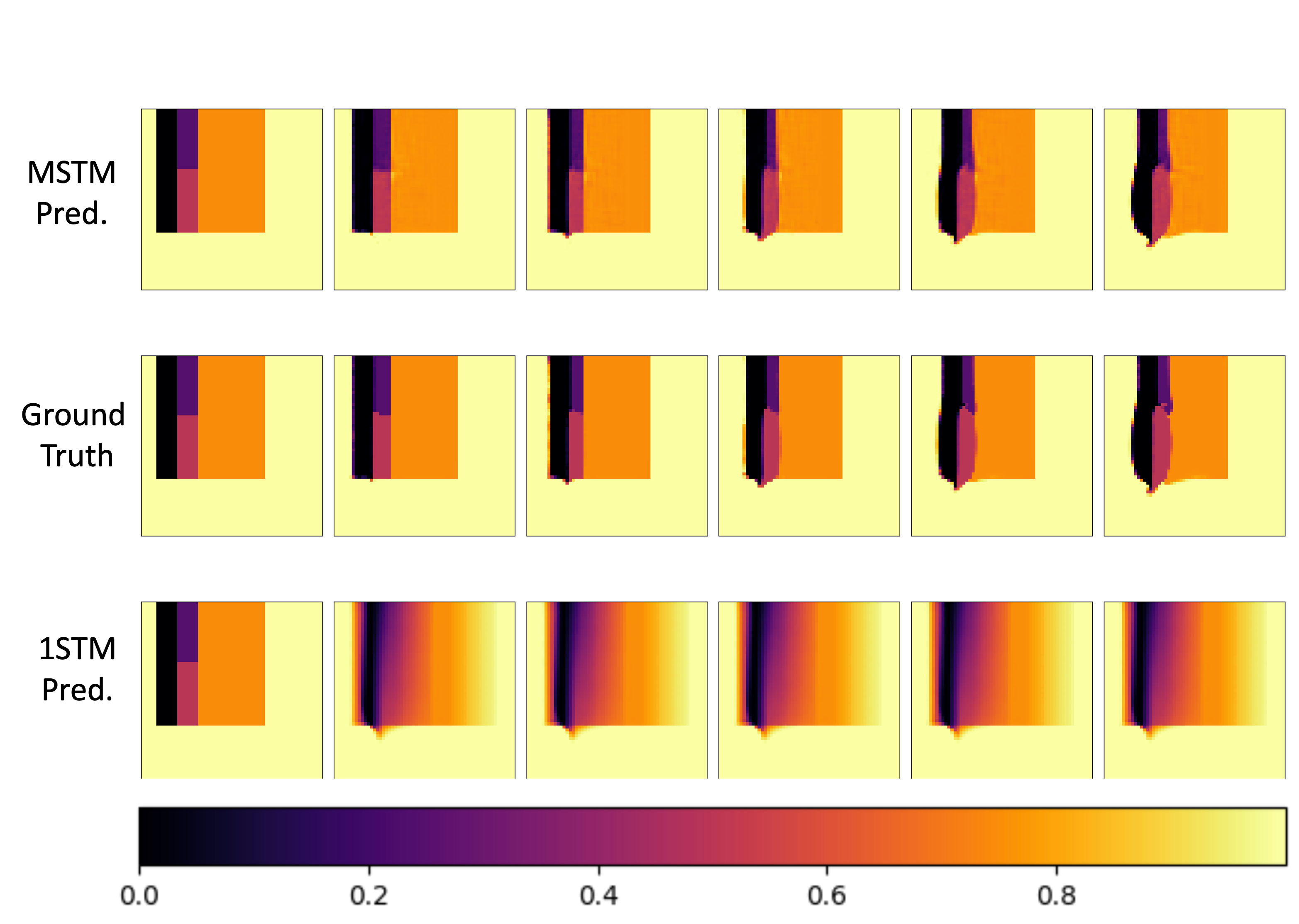}} &
    \subcaptionbox{Density\label{fig:app_density}}{%
      \includegraphics[width=0.40\textwidth]{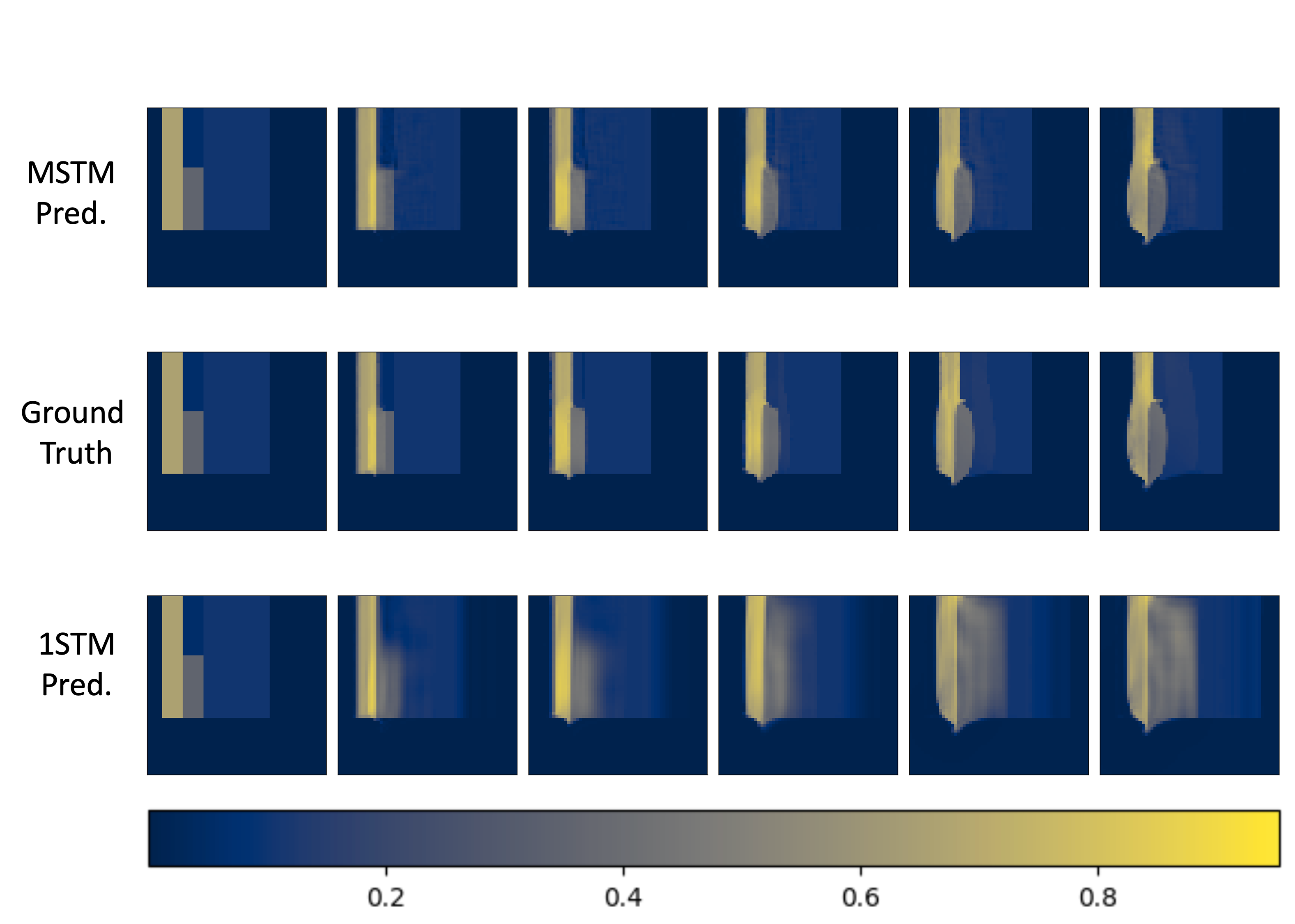}} \\[1ex]
    \subcaptionbox{Axial velocity $u_z$\label{fig:app_velocity_x}}{%
      \includegraphics[width=0.40\textwidth]{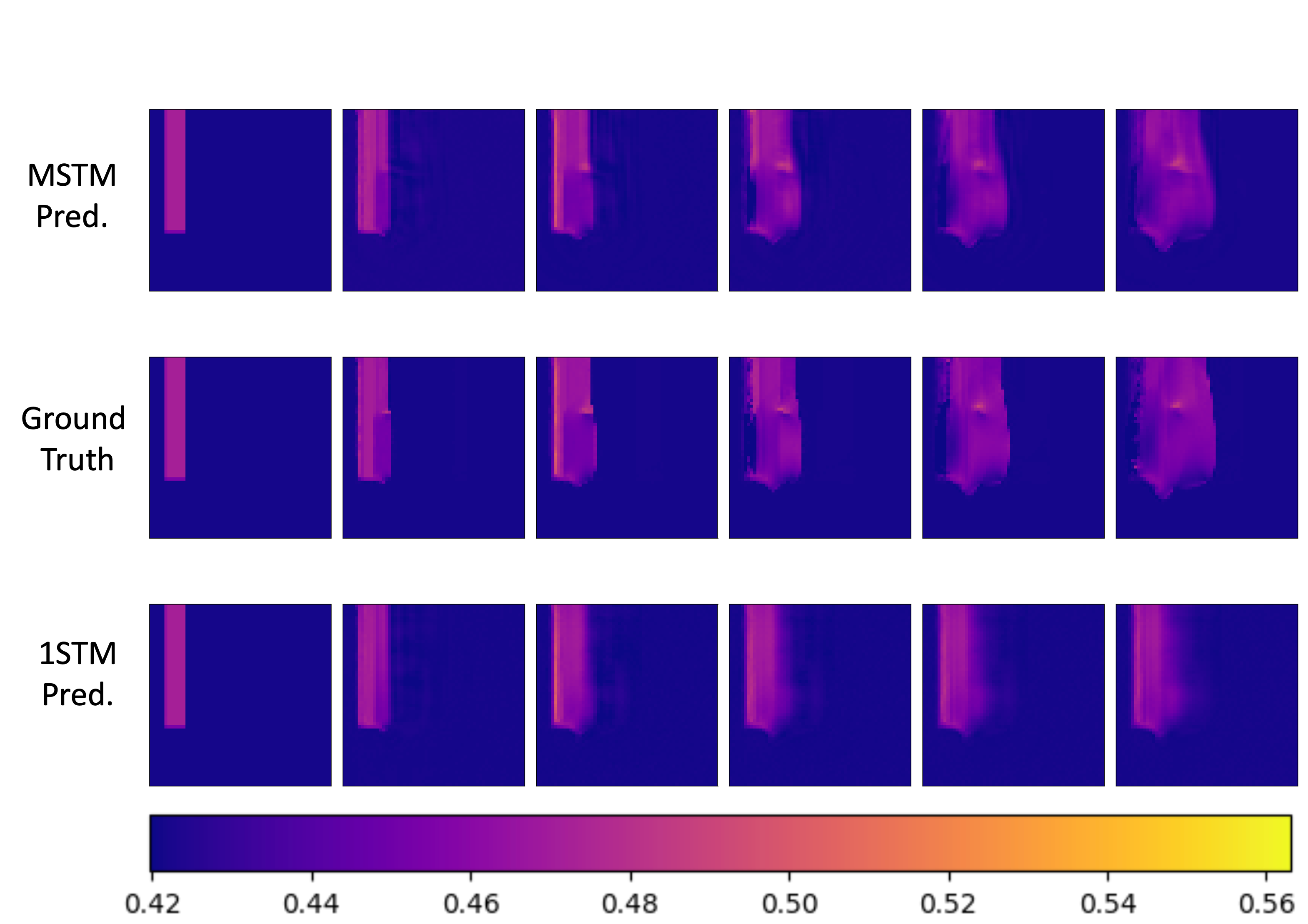}} &
    \subcaptionbox{Radial velocity $u_r$\label{fig:app_velocity_y}}{%
      \includegraphics[width=0.40\textwidth]{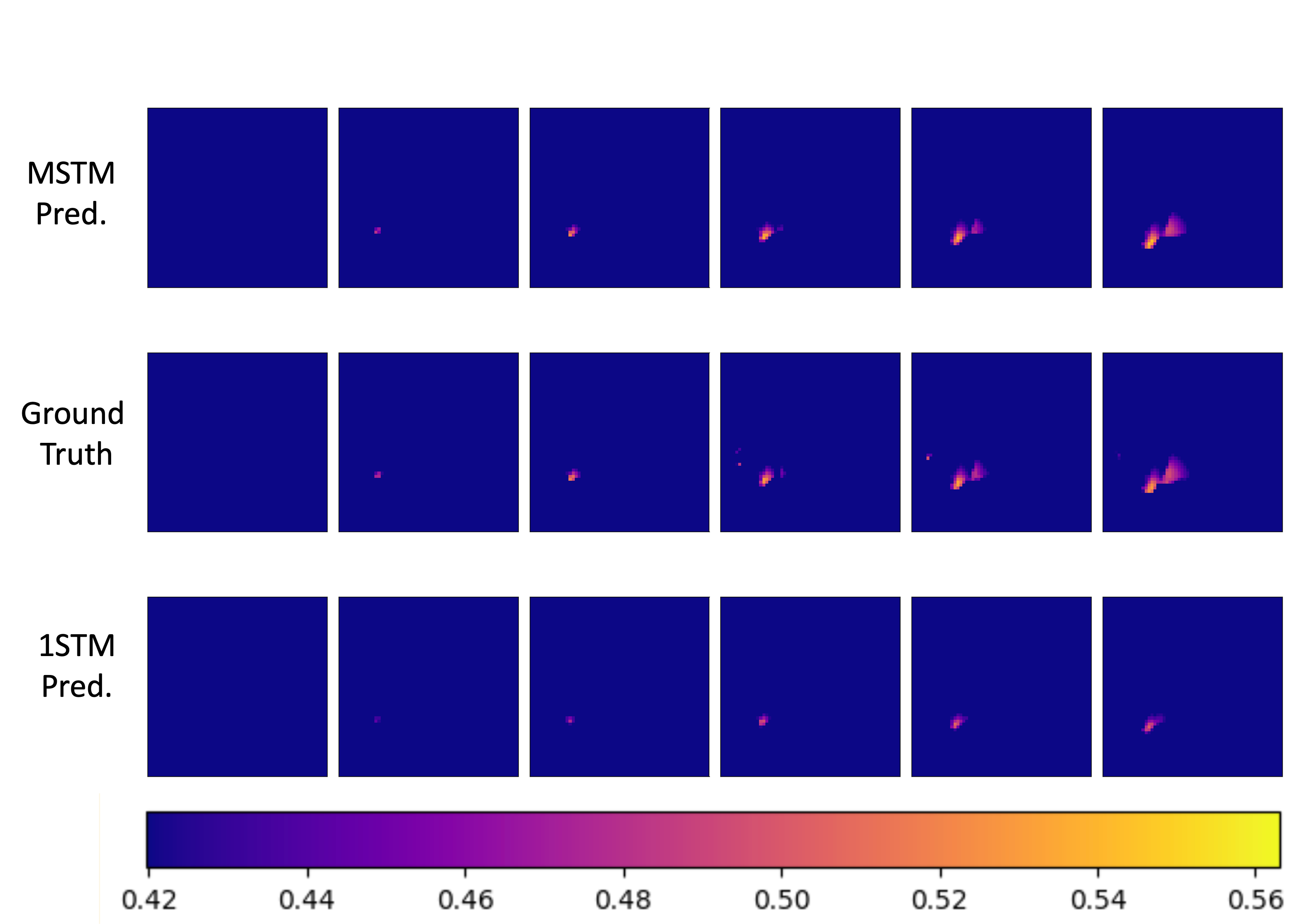}} \\[1ex]
    \subcaptionbox{Energy\label{fig:app_energy}}{%
      \includegraphics[width=0.40\textwidth]{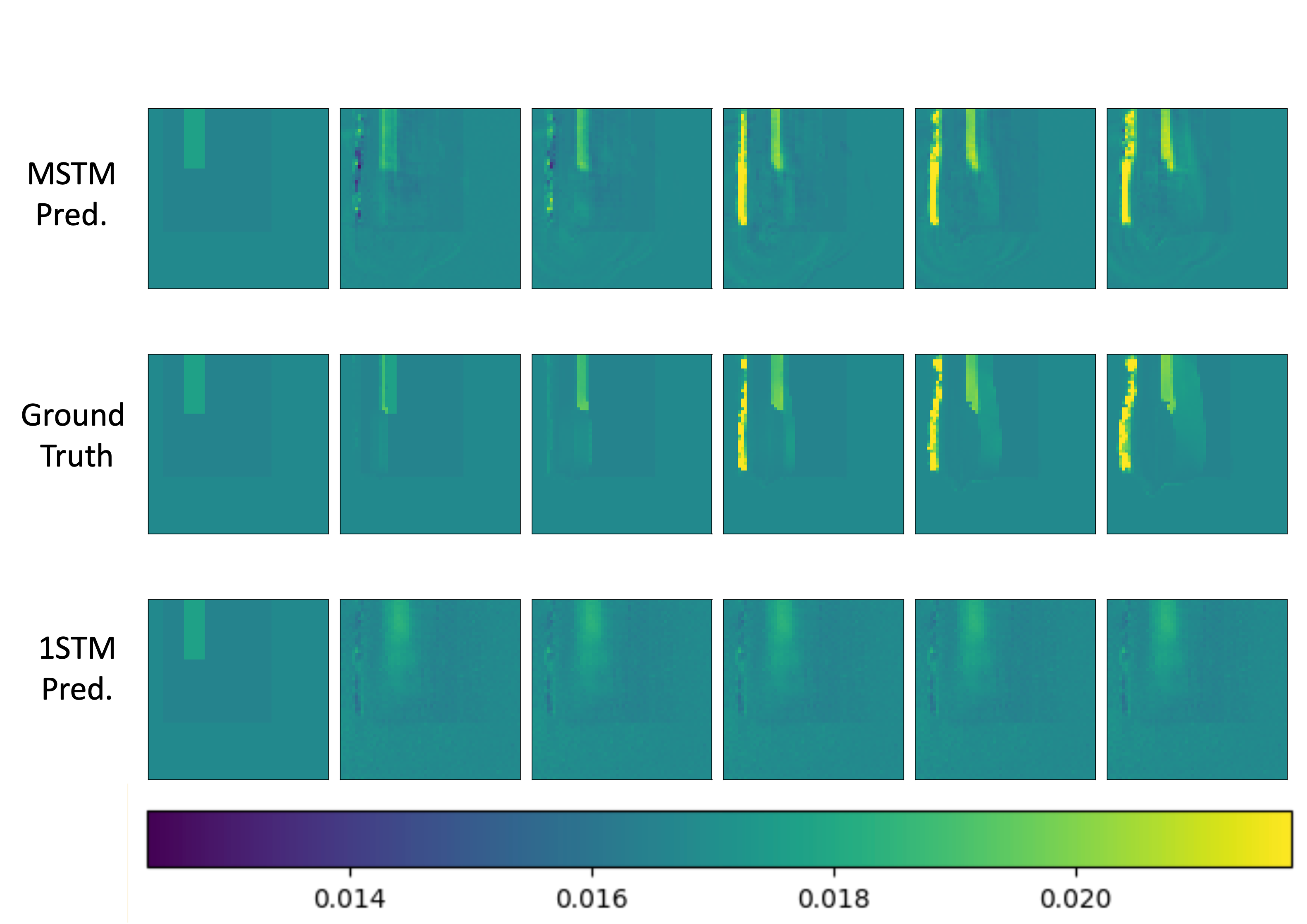}} &
    \subcaptionbox{Pressure\label{fig:app_pressure}}{%
      \includegraphics[width=0.40\textwidth]{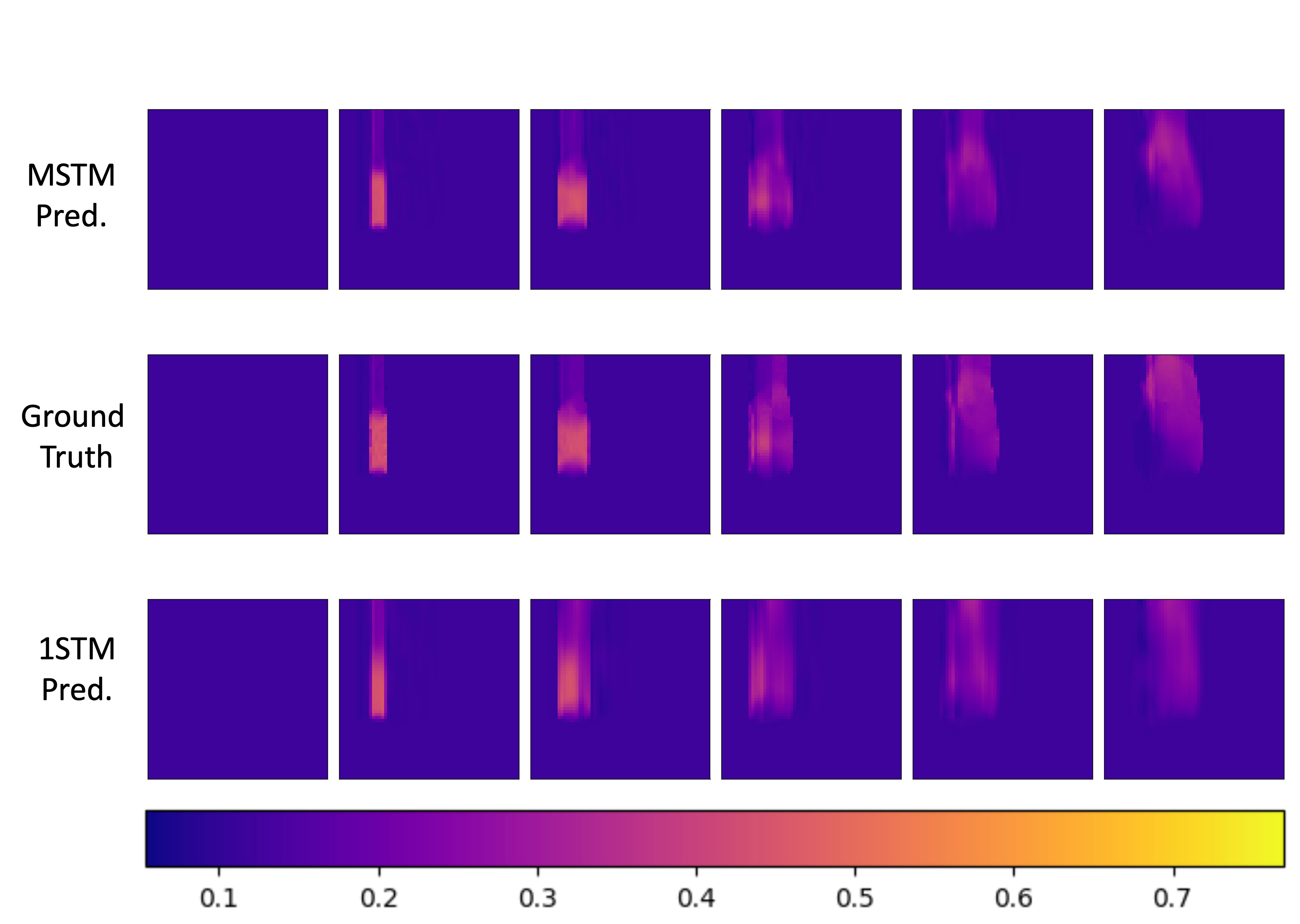}} \\[1ex]
    \multicolumn{2}{c}{%
      \subcaptionbox{Temperature\label{fig:app_temperature}}{%
        \includegraphics[width=0.40\textwidth]{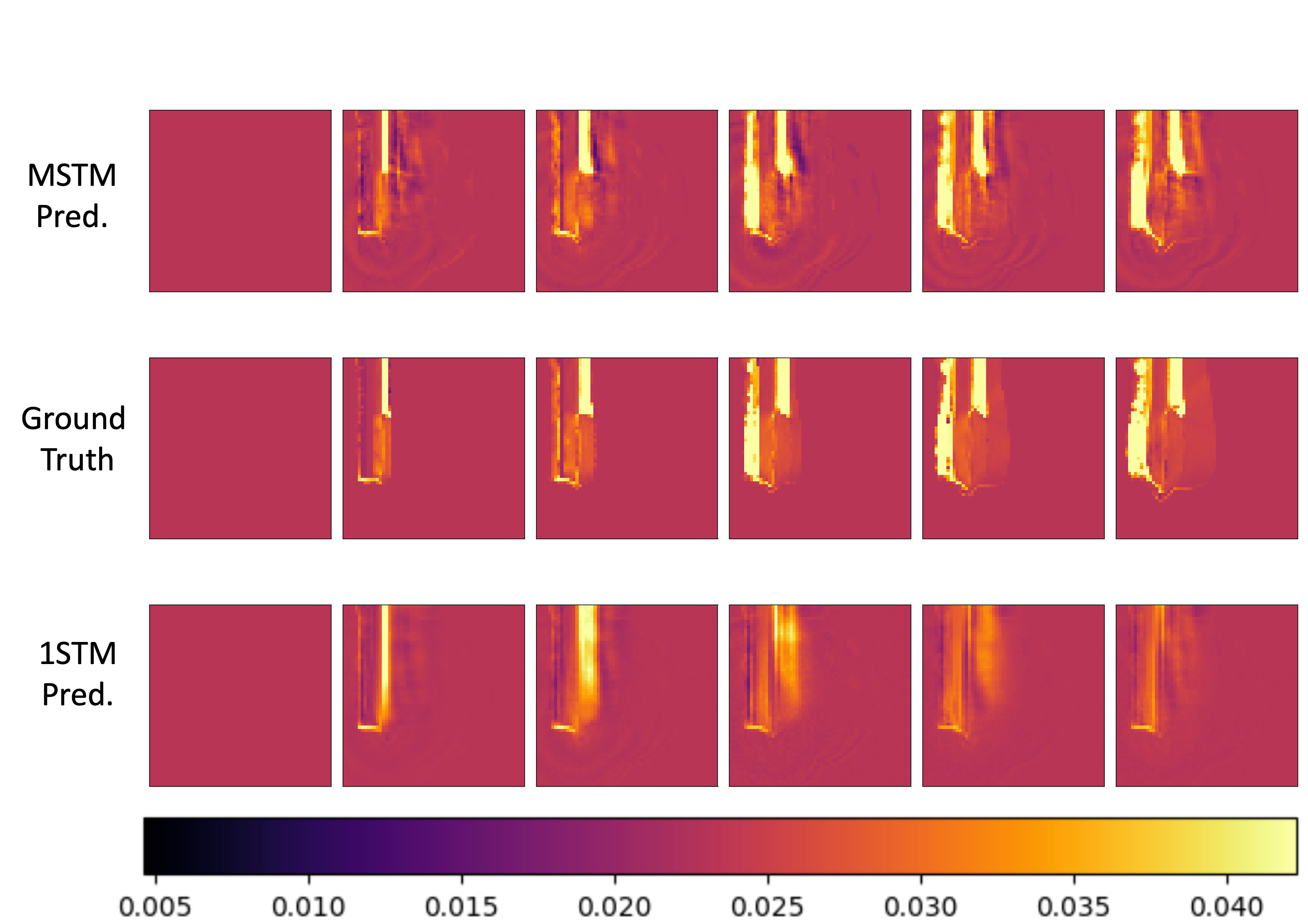}}}
  \end{tabular}
  \caption{Qualitative comparison for all seven fields. Each subfigure shows six evenly spaced time steps (0, 11, 22, 33, 44, 55). Within each panel, the \textbf{top row} is the MSTM prediction, the \textbf{middle row} is ground truth, and the \textbf{bottom row} is the 1STMs prediction. Color bars share the same limits across rows.}
  \label{fig:appendix_all_fields_m_vs_1}
\end{figure*}

Figure~\ref{fig:appendix_all_fields_m_vs_1} shows side-by-side sequences for all fields at six evenly spaced times. MSTM preserves shock fronts, pore-collapse interfaces, and hotspot localization across the rollout. Differences are most visible in the material indicator, pressure, and temperature. The predictions from the 1STMs smear interfaces and lag shock positions at later times, consistent with error accumulation under autoregression. In contrast, MSTM maintains sharper boundaries and coherent phase relations among density, pressure, and temperature, tracking the ground truth more closely.

\begin{figure*}[t]
\centering
\includegraphics[width=0.8\textwidth]{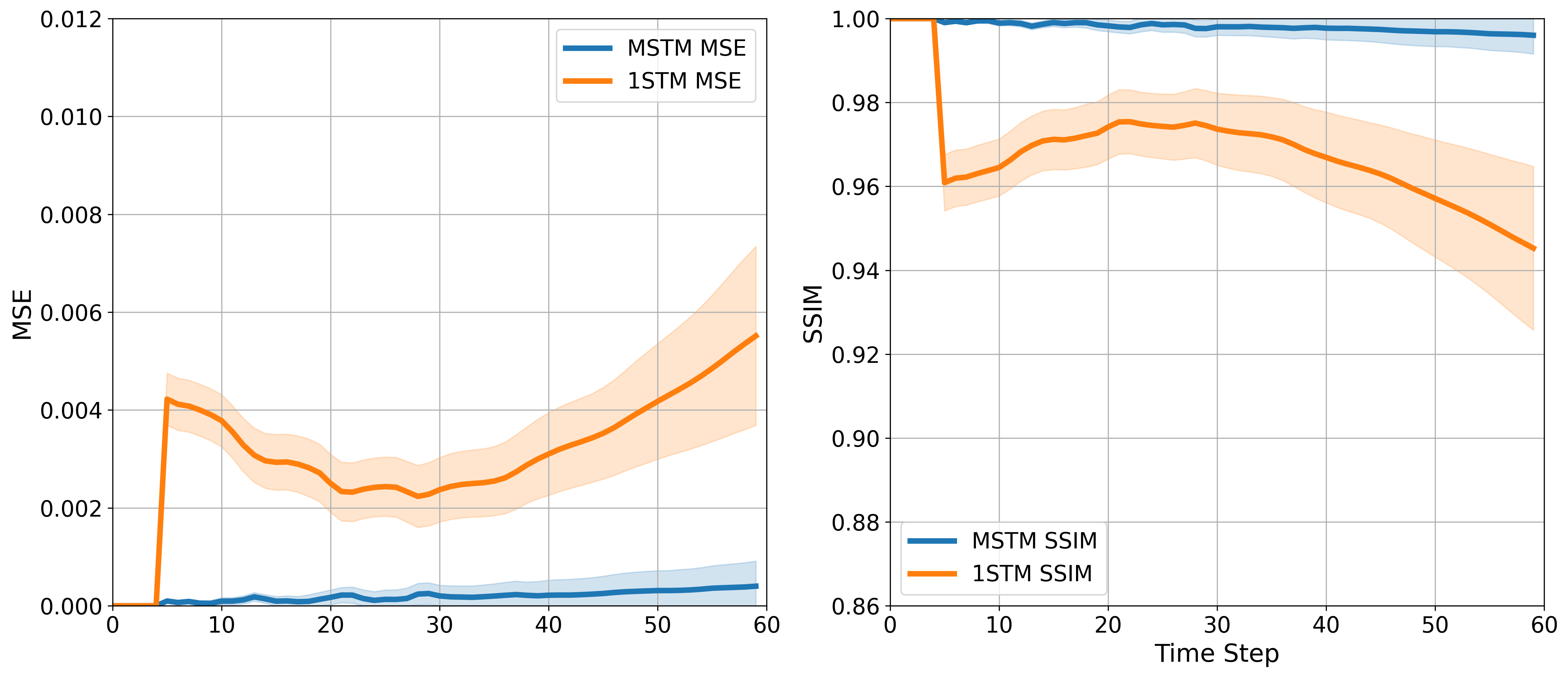}
\caption{Average MSE and SSIM over all fields comparing MSTM and 1STMs. Curves show means across test sequences; bands indicate one standard deviation.}
\label{fig:app_avg_mse_ssim}
\end{figure*}

Figure~\ref{fig:app_avg_mse_ssim} summarizes time-resolved averages across fields, with shaded bands showing one standard deviation over test cases. MSTM maintains low MSE and near-constant SSIM throughout the rollout. In contrast, the 1STM models show markedly higher MSE and lower SSIM. After an initial transient, the gap widens as autoregressive errors accumulate, especially at later times.

Three factors explain the multi-field advantage. First, shared spatial features align correlated structures such as shocks, rarefactions, and jetting that co-appear across fields. Second, updates are better timed because fields provide mutual cues: the material indicator and density inform pressure and temperature changes, while velocities constrain advection. Third, predicting all fields together reduces covariate shift, since each step uses a self-consistent seven-field state rather than mixing one predicted field with ground truth companions.

Aggregated over all test simulations and time steps, the 1STMs yield $\mathrm{MSE}=(3.1\pm0.6)\times10^{-3}$ and $\mathrm{SSIM}=0.968\pm0.009$, whereas MSTM attains $\mathrm{MSE}=(0.20\pm0.20)\times10^{-3}$ ($\approx$94\% lower) and $\mathrm{SSIM}=0.998\pm0.002$ ($\approx$94\% lower structural dissimilarity, i.e., $1-\mathrm{SSIM}$).

\subsection{Quantities of Interest}

\begin{figure*}[ht]
  \centering
  \subcaptionbox{Temperature QoI\label{fig:qoi_temperature}}{%
    \includegraphics[width=0.8\textwidth]{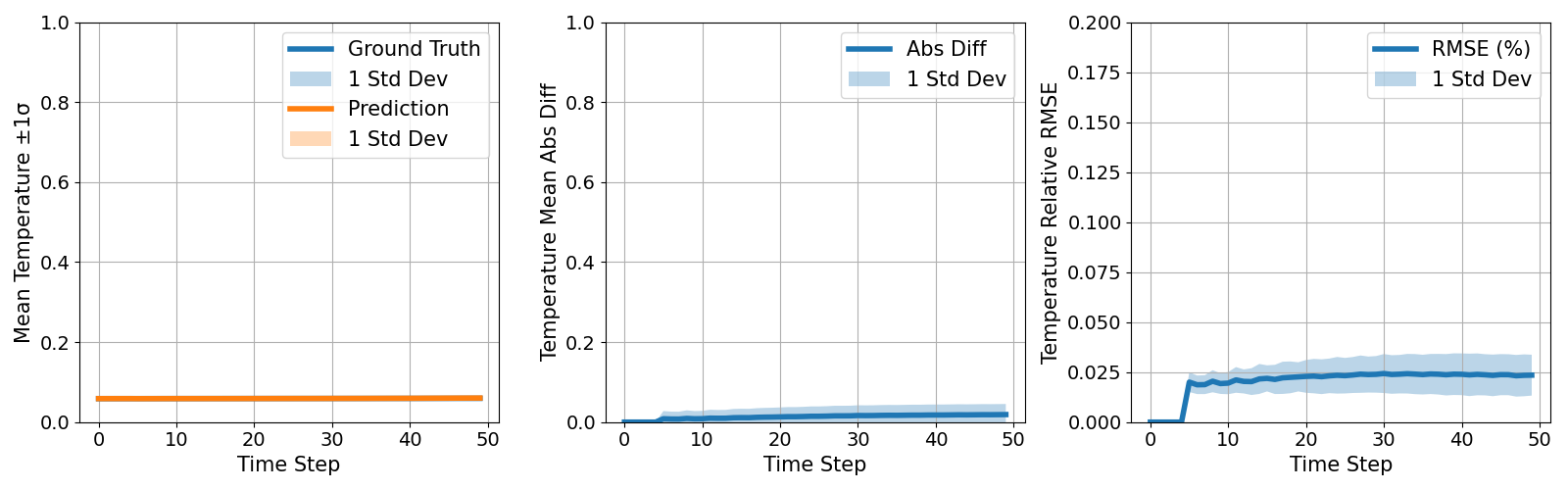}}\\[1ex]
  \subcaptionbox{Density QoI\label{fig:qoi_density}}{%
    \includegraphics[width=0.8\textwidth]{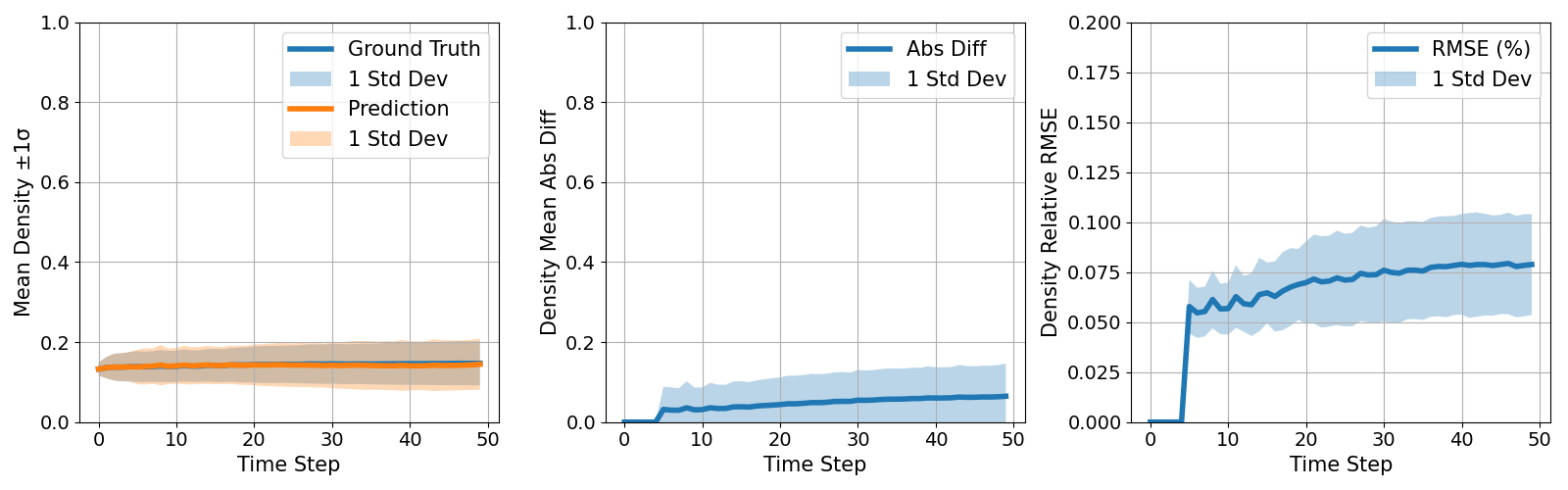}}\\[1ex]
  \subcaptionbox{Pressure QoI\label{fig:qoi_pressure}}{%
    \includegraphics[width=0.8\textwidth]{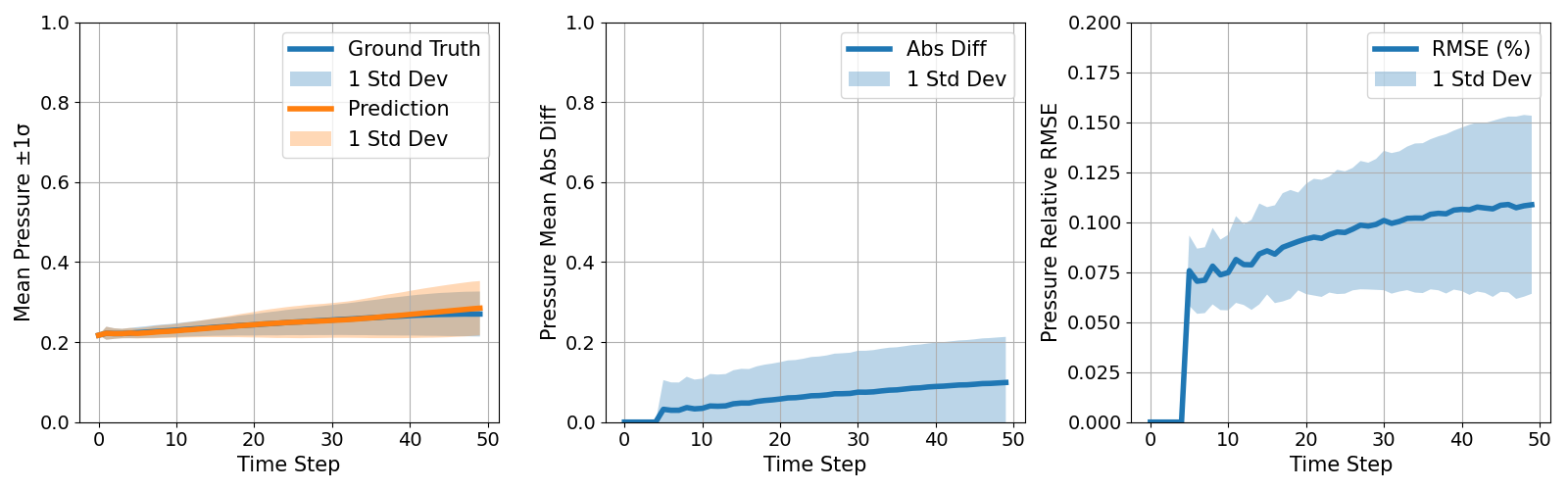}}\\
  \caption{Time series of quantities of interest for three fields in architected lattice simulations, masked to the material domain. Each row shows masked ground‑truth and predicted means $\pm1\sigma$ (left), masked mean absolute differences $\pm1\sigma$ (center) and masked relative RMSE $\pm1\sigma$ (right). MSTM recovers mean temperature, density and pressure within 5\% of ground truth over the rollout, with low variability across 325 test sequences.}
  \label{fig:qoi_metrics}
\end{figure*}

Beyond voxelwise metrics, we evaluate mass‑averaged quantities of interest (QoIs)—temperature, density and pressure—within the material domain. Figure~\ref{fig:qoi_metrics} plots time series of these QoIs for the most challenging lattice case: each row shows ground truth and predicted means (left), absolute differences (center) and relative RMSE (right), with $\pm1\sigma$ bands across 325 test runs. MSTM reproduces masked means within 5\% of ground truth and maintains small absolute and relative errors throughout the rollout. The low variance across sequences indicates stable performance across variations in porosity, lattice angle, and shock speed within the studied distribution.

These QoIs show that MSTM reproduces bulk thermodynamic quantities with good agreement, which is useful for reduced-cost design studies and parameter exploration. Rather than resolving the full simulation at each query, the surrogate learns a data-driven approximation of the effective dynamics and reproduces mass-averaged architected lattice temperature, density, and pressure within 5\% over the rollout.

To add a shock-specific diagnostic, we also compute a normalized shock-front location error ($\mathrm{SFE}_{\mathrm{norm}}$) from the pressure field. For each sample and time step, the two-dimensional pressure field is averaged over the transverse direction to obtain a one-dimensional axial profile, smoothed with a three-point moving average, and thresholded at $p_{\mathrm{thr}}(t)=p_{\mathrm{low}}(t)+0.5\left[p_{\max}(t)-p_{\mathrm{low}}(t)\right]$, where $p_{\mathrm{low}}(t)$ and $p_{\max}(t)$ are the 5th percentile and maximum of the ground truth axial profile at that same time step, respectively. The same ground truth-derived threshold is then applied to both the ground truth and predicted profiles, and the shock front is defined as the first axial position exceeding that threshold when searching from the impact side downstream. The fifth row of Figure~\ref{fig:metrics:combined} shows the resulting front-location error normalized by the 60-cell axial domain length. Fronts were detected for all saved snapshots in both datasets. Mean normalized errors remain small, at 0.0062 of the domain length for porous cases and 0.0102 for architected lattice cases, with larger late-time growth in the architected lattice case (0.0265 averaged over the last ten time steps versus 0.0088 for porous media), consistent with the higher sensitivity of thin struts and sharp interfaces to accumulated autoregressive errors.

\subsection{Comparison with Existing ML Surrogates}

Prior surrogate models for shock-dominated or extreme-condition simulations have often focused on reduced descriptors, single-field predictions, or problem settings that differ substantially in governing physics, geometry, and target outputs\cite{Moore2018,saad2024scalable}. In the present work, we therefore assess performance primarily through a controlled comparison between the proposed multi-field model and corresponding single-field spatio-temporal models trained and evaluated under identical normalization, rollout, and autoregressive settings. This comparison isolates the effect of joint multi-field prediction from confounding differences in architecture, training data, or evaluation protocol, which is important when comparing surrogate strategies within a single simulation framework\cite{Hunter2019}. Within that setting, joint multi-field prediction improves both accuracy and structural fidelity. These results support the use of coupled spatio-temporal surrogates for shock response in meso-structured materials, while broader comparisons across operator-learning, graph-based, or physics-constrained surrogate families remain an important direction for future work.

\section{Discussion} 
\label{sec:discussion}

Our results show that data-driven surrogates can reproduce the spatio-temporal evolution of shock waves in meso-structured materials at substantially lower cost than direct high-fidelity simulation. MSTM achieves mean squared error (MSE) as low as $2\times 10^{-4}$ on porous test cases and $1\times 10^{-3}$ on architected lattice cases, with structural similarity index (SSIM) above 0.98 and stable intersection-over-union (IoU) and mass-conservation (CM) errors throughout autoregressive rollouts. These metrics indicate that the network reproduces both local and global features while maintaining good agreement in phase-boundary location and low total-mass error over the rollout. A pressure-based shock-front diagnostic provides a more direct propagation measure, with mean normalized front-position errors of 0.0062 of the domain length for porous cases and 0.0102 for architected lattice cases. Differences between porous and architected lattice performance are physically interpretable: smooth porous interfaces are easier to predict, whereas thin struts and sharp edges in the architected lattice amplify sensitivity to small phase shifts during rollout. This is also reflected in the larger late-time architected lattice front error, which rises to 0.0265 of the domain length on average over the final ten time steps. Nevertheless, the model maintains strong accuracy across both configurations, achieving mean RMSE of 1.4\% in porous simulations and 3.2\% in architected lattice simulations. For the more challenging architected lattice cases, mass-averaged pressure, temperature, and density remain within about 5\% of the ground truth solution over the rollout.

A key result is that joint multi-field prediction improves accuracy and rollout stability relative to single-field surrogates trained under identical settings. In shock physics, pressure, density, temperature, and velocity are tightly coupled through the governing conservation laws and the equation of state. A single-field model must infer the evolution of sharp features from one channel alone, whereas MSTM can exploit correlated structures that co-evolve across fields, including shock fronts, rarefactions, and interface-driven features. In a controlled comparison, MSTM reduces MSE and structural dissimilarity $(1-\mathrm{SSIM})$ by about 94\% relative to seven single-field models (1STMs). The improved structural fidelity is most apparent near material interfaces and steep gradients, where autoregressive error accumulation can otherwise lead to interface smearing and phase misalignment.

The porous dataset includes strong dissipation associated with pore collapse and can exhibit non-monotonic shock response within the sampled parameter ranges. Within the evaluated distribution, MSTM reproduces these coupled thermodynamic and kinematic trends while maintaining stable autoregressive rollouts. The multi-field formulation also supports physically relevant summaries: although the architected lattice case is more challenging locally, some global integral quantities, such as the CM metric, remain slightly better preserved, consistent with the reduced sensitivity of integral measures to localized phase shifts. The more than three orders of magnitude reduction in inference time relative to direct simulation makes MSTM useful for repeated analysis and parameter studies, and it provides a pathway toward surrogate-assisted design workflows. Because the model evolves a full state rather than a single scalar objective, it can be coupled to downstream tasks that depend on spatial localization, interface evolution, and time-resolved thermodynamic history.

At the same time, the present validation remains bounded to the studied simulation setting, including the solver family, equations of state, dimensionality, grid representation, and parameter ranges considered here. The reported predictive performance therefore applies to variations in porosity, lattice angle, and loading conditions within that setting, rather than to arbitrary out-of-distribution shock problems. As initial conditions move farther outside the sampled range, for example to substantially higher shock speeds or geometries not represented in the training data, predictive reliability is expected to decrease because interface motion, compression rates, and thermodynamic response may depart from the patterns learned during training. This limitation is especially relevant for architected lattice configurations, where sharp geometric features and thin ligaments make accumulated autoregressive errors more visible over long rollouts.

Several extensions could broaden the applicability of MSTM. A first priority is to incorporate physics-informed objectives with conservation-aware constraints, rather than relying only on post hoc evaluation metrics to assess physical agreement. This may be especially valuable for the architected lattice cases, where autoregressive errors accumulate more visibly over long rollouts. Additional directions include uncertainty quantification for calibrated confidence estimates, extension of the multi-field autoregressive framework to higher-dimensional and more strongly coupled physics settings, and comparison with alternative surrogate families such as operator-learning, graph-based, or other physics-guided models. 

Taken together, these results show that coupled multi-field surrogates can serve as reduced-cost models for shock response in porous and architected lattice materials. In the present form, MSTM should be viewed as a practical surrogate for the class of problems studied here, with broader deployment depending on additional validation, benchmarking, and uncertainty-aware extensions.

\section*{Funding Statement}
This work performed under the auspices of the U.S. Department of Energy by Lawrence Livermore National Laboratory under Contract DE-AC52-07NA27344. This work was supported by the LLNL-LDRD Program under Project No. 24-ERD-036 and Project No. 24-ERD-005. The Lawrence Livermore National Security journal number is LLNL-JRNL-2011239.

\section*{Data Availability Statement}

The data and code associated with this study are subject to institutional and national laboratory restrictions and are therefore not publicly available.

\section*{Author Contributions} \label{contributions}

M.G.F.-G. conceived and led the study, developed the methodology, performed data curation and analysis, created figures and visualizations, and wrote the manuscript. M.H.S. designed and performed the hydrocode simulations, created the simulation setup figures, and ensured the accuracy of physics and simulation-related sections. W.J.S. contributed to the conception and design of the hydrocode simulations, provided supervision, and co-wrote the abstract, introduction, background, and conclusions. J.L. contributed to study conception, provided supervision, and made substantial revisions to the manuscript. J.L.B. reviewed the manuscript and provided strategic guidance. K.K. contributed to project discussions. M.K. provided feedback and funding support. All authors approved the final manuscript.

\bibliographystyle{elsarticle-num}
\bibliography{aipsamp}

\appendix

\section{Multi-field Temporal Model Architecture} \label{app:architecture}

The MSTM is designed for spatio-temporal prediction of shock dynamics in porous and architected lattice materials. It utilizes a hybrid architecture that combines convolutional layers for spatial feature extraction with an LSTM module to capture temporal dependencies within the input data. The model consists of 36,834,032 trainable parameters. It operates on input data of shape \((\text{batch\_size}, T, 7, 60, 60)\), where \(T\) is the temporal window size, and each sample consists of seven physical fields with a spatial resolution of \(60 \times 60\) (pressure, density, material indicator, energy, temperature, and two in-plane velocity components). These correspond to velocity in \(r\) and \(z\) for the porous case, and velocity in \(x\) and \(y\) for the architected lattice case. The data is normalized independently for each field to the interval \([0,1]\) using min-max scaling. The minimum and maximum values are computed across all time steps and training sequences, ensuring a consistent transformation. The same normalization parameters are applied to the validation and test datasets.

The model is trained autoregressively to predict the next time step from the previous $T$ frames. It outputs $(\text{batch\_size},1,7,60,60)$ and is rolled out by sliding the input window forward and feeding predictions back as inputs. The CNN module extracts hierarchical spatial features at each time step, which are then processed by the LSTM module to learn the temporal evolution of shock propagation. The final fully connected layer reconstructs the output at each time step, ensuring consistency with the underlying physical system.

Table \ref{table:architecture.} provides a detailed description of the model architecture, training setup, and implementation details to facilitate reproducibility.

\begin{table*}
    \centering
    \caption{Detailed specification of the MSTM network architecture. The convolutional feature extractor comprises two Conv2d layers:  
\textbf{conv1} applies 64 filters of size 3$\times$3 (\text{padding}=1) to the seven-field input, followed by ReLU activation and 2$\times$2 max pooling;  
\textbf{conv2} uses 128 filters of size 3$\times$3 (\text{padding}=1), ReLU activation, and 2$\times$2 max pooling. 
The LSTM module accepts the flattened feature maps (128$\times$15$\times$15) and processes them through a four-layer stacked LSTM with hidden dimension 512, initialized with zero states \((h_{0},c_{0})\). 
A fully connected layer then projects the final LSTM hidden state (512 units) to a vector of length \(7\times60\times60\), which is reshaped to \((\text{batch\_size},1,7,60,60)\) to produce the next-step field prediction.
}
    \begin{tabular}{ll}
        \toprule
        \textbf{Component} & \textbf{Configuration} \\
        \midrule
        \textbf{Convolutional Feature Extractor} & \\
        conv1 & \(\text{Conv2d}(7, 64, 3, \text{padding}=1)\) followed by ReLU activation and MaxPool2d(2, 2) \\
        conv2 & \(\text{Conv2d}(64, 128, 3, \text{padding}=1)\) followed by ReLU activation and MaxPool2d(2, 2) \\
        \midrule
        \textbf{LSTM Module} & \\
        lstm & \(\text{LSTM}(128 \times 15 \times 15, 512, 4, \text{batch\_first}=True)\) \\
        Hidden state & \( h_0, c_0 = \text{Zeros}(4, \text{batch\_size}, 512) \) \\
        \midrule
        \textbf{Fully Connected Layer} & \\
        fc & \(\text{Linear}(512, 7 \times 60 \times 60)\) \\
        Output Reshape & \(\text{Reshape}(\text{batch\_size}, 1, 7, 60, 60)\) \\
        \bottomrule
    \end{tabular}
    \label{table:architecture.}
\end{table*}

The model is trained using the Adam optimizer with a learning rate of \(0.0005\) and the mean squared error (MSE) loss function. Training is conducted for 1000 epochs with a \text{batch\_size} of 256.

\section{Performance Metrics}\label{app:metrics}

This appendix states the exact formulas used to compute the five performance metrics: mean squared error (MSE), intersection over union (IoU), structural similarity index measure (SSIM), conservation of mass (CM), and normalized shock-front location error ($\mathrm{SFE}_{\mathrm{norm}}$). Let $p_f^{(s)}(i,j;t)$ and $g_f^{(s)}(i,j;t)$ denote the predicted and ground truth values, respectively, for field $f \in \{1,\dots,7\}$ at grid index $(i,j)$, time $t$, and sample $s$. Sums run over all grid points of the computational mask $M^{(s)}(t)$ unless stated otherwise. When results are reported as a single number, we average over fields, time steps, and test samples.

\subsection{Mean Squared Error (MSE)}
For a single field and time step,
\begin{equation}
\mathrm{MSE}_f^{(s)}(t)
=\frac{1}{|M^{(s)}(t)|}\sum_{(i,j)\in M^{(s)}(t)}\!\!
\bigl(p_f^{(s)}(i,j;t)-g_f^{(s)}(i,j;t)\bigr)^2.
\end{equation}
We report $\mathrm{MSE}$ averaged over $f$, $t$, and $s$.

\subsection{Intersection over Union (IoU) on the material indicator field}
IoU is computed on the material indicator field only. Because this field is continuous, we use a soft interval membership to define overlap. IoU is computed over the full $H\times W$ grid (no additional spatial mask), since material membership is already defined through the soft indicators $m_p$ and $m_g$. Let $p^{(s)}(i,j;t)$ and $g^{(s)}(i,j;t)$ be the predicted and true material values for sample $s$, and let $[l_b,u_b]$ be the interval that corresponds to \textit{material} (as opposed to void). We define smooth memberships
\begin{align}
m_p(i,j;t) &= \sigma\!\left(\frac{p^{(s)}(i,j;t)-l_b}{k}\right)\,
              \sigma\!\left(\frac{u_b-p^{(s)}(i,j;t)}{k}\right),\\
m_g(i,j;t) &= \sigma\!\left(\frac{g^{(s)}(i,j;t)-l_b}{k}\right)\,
              \sigma\!\left(\frac{u_b-g^{(s)}(i,j;t)}{k}\right),
\end{align}
where $\sigma(x)=1/(1+e^{-x})$ and $k$ controls the smoothness near the bounds (we use $k=0.05(u_b-l_b)$). The soft IoU is then
\begin{equation}
\mathrm{IoU}^{(s)}(t)
=\frac{\sum_{i=1}^{H}\sum_{j=1}^{W} \min\!\bigl(m_p(i,j;t),\,m_g(i,j;t)\bigr)}
       {\sum_{i=1}^{H}\sum_{j=1}^{W} \max\!\bigl(m_p(i,j;t),\,m_g(i,j;t)\bigr)}.
\end{equation}
Bounds differ by dataset and reflect how the material indicator is represented:
\[
\text{porous: } [l_b,u_b]=[0.20,\,0.30], \qquad
\text{lattice: } [l_b,u_b]=[0.54,\,0.99].
\]

\subsection{Structural Similarity Index Measure (SSIM)}
We compute SSIM for each field using the standard definition:
\begin{equation}
\mathrm{SSIM}(p_f,g_f)=
\frac{(2\mu_{p}\mu_{g}+C_1)(2\sigma_{pg}+C_2)}
     {(\mu_{p}^{2}+\mu_{g}^{2}+C_1)(\sigma_{p}^{2}+\sigma_{g}^{2}+C_2)},
\end{equation}
where $\mu_{p},\mu_{g}$ are spatial means, $\sigma_{p}^{2},\sigma_{g}^{2}$ are variances, $\sigma_{pg}$ is covariance, and $C_1,C_2$ are stabilization constants. We report SSIM averaged over fields, time, and samples.

\subsection{Conservation of Mass (CM)}
Total mass for a snapshot is obtained by summing density, optionally weighted by a material indicator field $f(i,j;t)$ when present:
\begin{equation}
\begin{split}
M_p^{(s)}(t) &= \sum_{(i,j)} \rho_p^{(s)}(i,j;t)\,f_p^{(s)}(i,j;t), \\
M_g^{(s)}(t) &= \sum_{(i,j)} \rho_g^{(s)}(i,j;t)\,f_g^{(s)}(i,j;t).
\end{split}
\end{equation}
If $f$ is not used, set $f\equiv 1$. The reported CM is the relative mass error,
\begin{equation}
\mathrm{CM}^{(s)}(t)=\frac{\bigl|M_p^{(s)}(t)-M_g^{(s)}(t)\bigr|}
                          {M_g^{(s)}(t)}.
\end{equation}
We average $\mathrm{CM}^{(s)}(t)$ over time and samples.

\subsection{Normalized Shock-Front Location Error ($\mathrm{SFE}_{\mathrm{norm}}$)}
To obtain a shock-oriented propagation diagnostic, we extract the front from the pressure field. For each sample and time step, the pressure field is averaged over the transverse direction to form an axial profile,
\begin{equation}
\bar{p}^{(s)}(x;t)=\frac{1}{N_{\perp}}\sum_{y=1}^{N_{\perp}} p_{\mathrm{press}}^{(s)}(x,y;t), \qquad
\bar{g}^{(s)}(x;t)=\frac{1}{N_{\perp}}\sum_{y=1}^{N_{\perp}} g_{\mathrm{press}}^{(s)}(x,y;t),
\end{equation}
which is then smoothed with a three-point moving average. We define a ground truth-based threshold
\begin{equation}
p_{\mathrm{thr}}^{(s)}(t)=p_{\mathrm{low}}^{(s)}(t)+\alpha\left(p_{\max}^{(s)}(t)-p_{\mathrm{low}}^{(s)}(t)\right), \qquad \alpha=0.5,
\end{equation}
where $p_{\mathrm{low}}^{(s)}(t)$ is the 5th percentile of the current ground truth axial profile $\bar{g}^{(s)}(x;t)$ and $p_{\max}^{(s)}(t)=\max_x \bar{g}^{(s)}(x;t)$. This threshold is computed from the ground truth profile only and then applied unchanged to both the ground truth and predicted profiles. The true and predicted shock-front positions are then
\begin{equation}
x_{\mathrm{front,true}}^{(s)}(t)=\min\left\{x:\bar{g}^{(s)}(x;t)\ge p_{\mathrm{thr}}^{(s)}(t)\right\},
\qquad
x_{\mathrm{front,pred}}^{(s)}(t)=\min\left\{x:\bar{p}^{(s)}(x;t)\ge p_{\mathrm{thr}}^{(s)}(t)\right\},
\end{equation}
searching from the impact side downstream. The front-position error is first computed in grid-cell units as
\begin{equation}
\Delta x_{\mathrm{front}}^{(s)}(t)=\left|x_{\mathrm{front,pred}}^{(s)}(t)-x_{\mathrm{front,true}}^{(s)}(t)\right|.
\end{equation}
The reported shock-front location error is then normalized by the axial domain length,
\begin{equation}
\mathrm{SFE}_{\mathrm{norm}}^{(s)}(t)=\frac{\Delta x_{\mathrm{front}}^{(s)}(t)}{N_x}, \qquad N_x=60.
\end{equation}
Because the saved rollout tensors are normalized, thresholding is applied consistently in normalized pressure space, while $\Delta x_{\mathrm{front}}^{(s)}(t)$ is measured in grid cells and then divided by the 60-cell axial domain length to yield dimensionless $\mathrm{SFE}_{\mathrm{norm}}^{(s)}(t)$.

The reported metric values are computed on the normalized fields used by the network, with each field scaled to $[0,1]$ using training-set extrema. The $\mathrm{SFE}_{\mathrm{norm}}$ metric uses the normalized pressure field for thresholding and reports the resulting front-position error as a normalized fraction of the 60-cell axial domain length.

\subsection{Averaging conventions}
Let $\langle \cdot \rangle$ denote averaging over fields (when applicable), time steps, and test samples. For each metric $\mathcal{Q}\in\{\mathrm{MSE},\mathrm{IoU},\mathrm{SSIM},\mathrm{CM},\mathrm{SFE}_{\mathrm{norm}}\}$, we report $\langle \mathcal{Q}\rangle$ as the mean and its standard deviation across the evaluation set.

\section{Governing Equations and Physical Modeling} \label{sec:gov_eqs}

The dynamics of shock propagation in porous media and architected lattice materials are governed by the fundamental conservation laws of mass, momentum, and energy. These equations describe the behavior of compressible fluids and are numerically solved in this study using LLNL's MARBL hydrocode. The governing equations are expressed as:

\begin{align}
    \frac{\partial \rho}{\partial t} + \nabla \cdot (\rho \mathbf{u}) &= 0, \label{eq:mass} \\
    \frac{\partial (\rho \mathbf{u})}{\partial t} + \nabla \cdot (\rho \mathbf{u} \mathbf{u} + p \mathbf{I}) &= 0, \label{eq:momentum} \\
    \frac{\partial E}{\partial t} + \nabla \cdot [(E + p) \mathbf{u}] &= 0. \label{eq:energy}
\end{align}

\noindent Here, $\rho$ is the density, $\mathbf{u}$ is the velocity vector, $p$ is the pressure, $\mathbf{I}$ is the identity tensor, and $E$ is the total energy per unit volume, given by:

\begin{equation}
    E = \rho e + \frac{1}{2} \rho \mathbf{u}^2,
\end{equation}

\noindent where $e$ is the specific internal energy. The pressure $p$ is determined from an equation of state (EOS), which relates pressure, density, and internal energy. Livermore equation-of-state (LEOS) tables \cite{barrios2012precision} were used to model air (LEOS 2260), aluminum (LEOS 130), lithium fluoride (LEOS 2240), copper (LEOS 290), tantalum (LEOS 730), and tungsten (LEOS 740). For strength modeling, Steinberg-Guinan models were used for some materials, elastic linear-plastic models were used for others, and strength was turned off for air.

For the porous setup, porosity is dynamically modeled using a homogenized approach, where material properties such as density and compressibility scale with porosity $\phi$:

\begin{equation}
    \phi = 1 - \frac{\rho}{\rho_{\text{solid}}},
\end{equation}

\noindent where $\rho_{\text{solid}}$ represents the density of the fully dense material. The density satisfies the constraint

\begin{equation}
    0 \leq \rho \leq \rho_{\text{solid}},
\end{equation}

\noindent with $\rho = 0$ corresponding to the absence of material and $\rho = \rho_{\text{solid}}$ to the fully dense state. To capture the material response in the small-pore and strong-shock regime, under-dense initialization with a pressure floor of zero is used. The porosity evolves dynamically under shock loading, allowing for accurate representation of compaction effects.

\section{Numerical Implementation} \label{sec:num_methods}

The numerical solution of the governing equations is implemented using MARBL, a high-order multiphysics simulation code developed at Lawrence Livermore National Laboratory. MARBL employs a finite element discretization scheme together with an artificial hyperviscosity term that provides robust shock-capturing capability. 

\subsection{Discretization and Time Integration}

The computational domain is discretized using finite elements with significant resolution in certain areas, such as the porous layers and material interfaces. A high-order shock-capturing scheme \cite{bello2020matrix} is employed to resolve steep gradients and discontinuities at shock fronts while maintaining numerical stability.

Temporal integration is performed using a second-order explicit Runge-Kutta like scheme. An adaptive time-stepping approach is utilized, where the time step is determined by the Courant–Friedrichs–Lewy (CFL) condition:

\begin{equation}
    \Delta t = \text{CFL} \cdot \min \left( \frac{\Delta x}{|\mathbf{u}| + c} \right),
\end{equation}

\noindent where $\Delta x$ is the grid spacing, $c$ is the local sound speed, and CFL is a stability factor less than 1.

\subsection{Boundary Conditions and Computational Efficiency}

For the porous configuration, all external boundaries use no-flux conditions and cylindrical symmetry is imposed about the central axis, so mass, momentum, and energy remain confined to the domain while capturing axisymmetric physics. For the architected lattice configuration, no-flux conditions are applied on all boundaries except the left, which is treated as a free-flow boundary to admit the impact-driven wave. The architected lattice is modeled in 2D Cartesian geometry, which reduces computational cost while preserving the essential three-dimensional behavior.

MARBL is optimized for high-performance computing (HPC) environments, leveraging parallel computing to efficiently simulate large-scale shock interactions in porous and architected lattice materials. This capability allows for high-resolution studies of wave dynamics, rarefaction effects, and energy dissipation mechanisms across a wide range of initial conditions.

\section{Mass-Averaged Field Error Analysis in Shocked Architected Lattice Materials} \label{app:qoi_metrics}

\subsection{Data Preparation}

We begin by reshaping the architected lattice simulation raw test tensor so that its dimensions correspond to
\[
  (N,\;T,\;L,\;F,\;H,\;W)
  =
  (N,\;50,\;5,\;7,\;60,\;60),
\]
\noindent where:
\begin{itemize}[noitemsep]
  \item $N$ is the number of test samples.
  \item $T=50$ is the number of time steps per sample.
  \item $L=5$ is the auto‐regressive chain length.
  \item $F=7$ is the number of physical fields.
  \item $H=W=60$ are the spatial grid dimensions.
\end{itemize}

We use lower-case indices $s\in{1,\dots,N}$, $t\in{1,\dots,T}$, $\ell\in{1,\dots,L}$, $f\in{1,\dots,F}$, $i\in{1,\dots,H}$, and $j\in{1,\dots,W}$ to denote the sample, time step, window position, field type, and spatial coordinates, respectively.

\subsection{Material Indicator Masking}

The fourth field (index 3) encodes a normalized material indicator, $m\in[0,1]$. We denote the ground truth and predicted indicators by
\[
m_{\mathrm{g}}(s,t,\ell,i,j)=X_{\mathrm{g}}(s,t,\ell,3,i,j),
\qquad
m_{\mathrm{p}}(s,t,\ell,i,j)=X_{\mathrm{p}}(s,t,\ell,3,i,j),
\qquad
m_{\mathrm{g}},m_{\mathrm{p}}\in[0,1].
\]
Here $s\in\{1,\dots,N\}$, $t\in\{1,\dots,T\}$, $\ell\in\{1,\dots,L\}$, and $(i,j)\in\{1,\dots,H\}\times\{1,\dots,W\}$. The subscript $\mathrm{g}$ denotes \textit{ground truth}, while the subscript $\mathrm{p}$ denotes \textit{prediction}.

We define lower and upper thresholds based on the initial material value assignment,
\[
  \mathrm{lb}=0.54,\quad \mathrm{ub}=0.99,
\]
This mask selects the contiguous material domain, here the architected aluminum lattice. The ground truth-based mask is constructed as follows:
\begin{align}
M_{\mathrm{g}}(s,t,\ell,i,j) &=
\begin{cases}
1, & \text{if } \mathrm{lb} \le m_{\mathrm{g}}(s,t,\ell,i,j) \le \mathrm{ub},\\
0, & \text{otherwise}.
\end{cases}
\end{align}
and similarly for the prediction-based mask:
\begin{align}
M_{\mathrm{p}}(s,t,\ell,i,j) &=
\begin{cases}
1, & \text{if } \mathrm{lb} \le m_{\mathrm{p}}(s,t,\ell,i,j) \le \mathrm{ub},\\
0, & \text{otherwise}.
\end{cases}
\end{align}

The masked ground truth and prediction for field $f$ are
\begin{align}
g_{f}(s,t,\ell,i,j) &= X_{\mathrm{g}}(s,t,\ell,f,i,j)\cdot M_{\mathrm{g}}(s,t,\ell,i,j),\\
p_{f}(s,t,\ell,i,j) &= X_{\mathrm{p}}(s,t,\ell,f,i,j)\cdot M_{\mathrm{p}}(s,t,\ell,i,j).
\end{align}

\begin{figure*}[!htb]
  \centering
  \begin{subfigure}[b]{0.4\textwidth}
    \includegraphics[width=\textwidth]{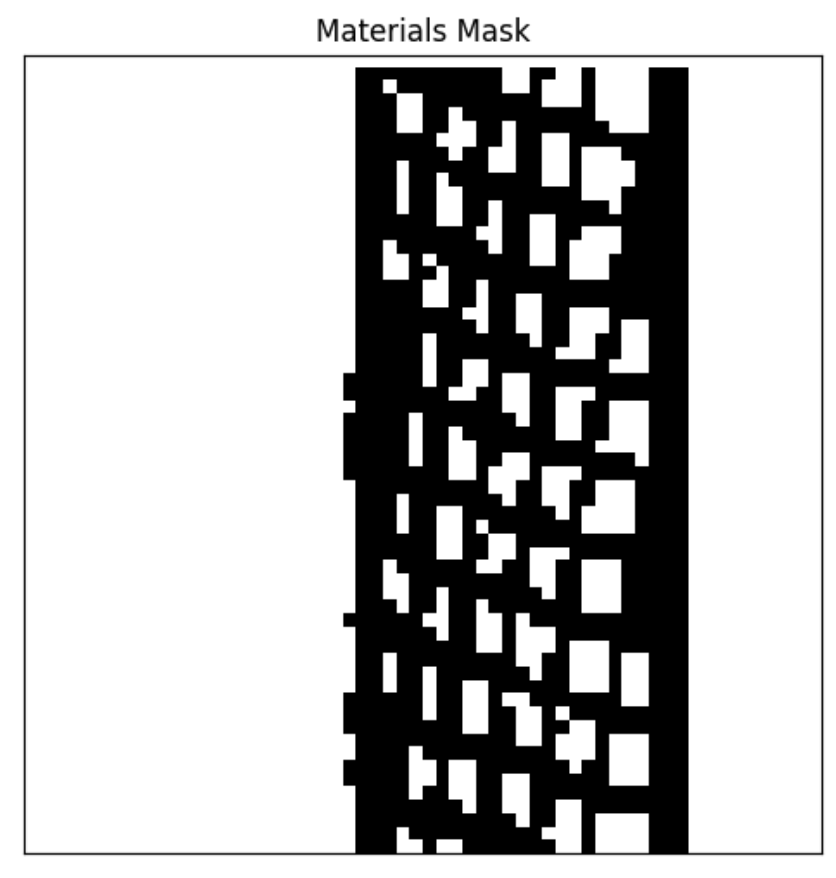}
    \caption{$M_{\mathrm{g}} (t=0)$.}
    \label{fig:mask_t0}
  \end{subfigure}
  \begin{subfigure}[b]{0.4\textwidth}
    \includegraphics[width=\textwidth]{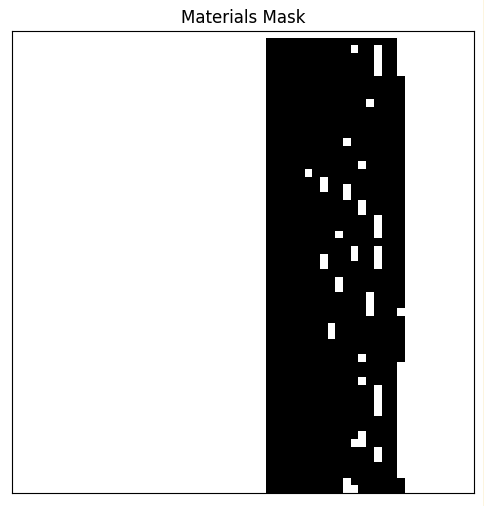}
    \caption{$M_{\mathrm{g}}(t=49)$.}
    \label{fig:mask_t49}
  \end{subfigure}
  \caption{Evolution of the material indicator mask. These plots derive from ground truth data; no machine learning methods were applied.}
  \label{fig:material_masks}
\end{figure*}

Figure \ref{fig:material_masks} shows the mask at the initial and final time steps. Black pixels satisfy $\mathrm{lb}\le m\le \mathrm{ub}$.

The absolute difference field is defined as
\begin{align}
d_{f}(s,t,\ell,i,j) &= |g_{f}(s,t,\ell,i,j) - p_{f}(s,t,\ell,i,j)|.
\end{align}

\subsection{Aggregation Across Samples}

Before diving into the set definitions, we simplify the notation by dropping the explicit sample index \(s\) and the spatial indices \((i,j)\). Instead of writing out every 
\[
g_f^{(s)}(i,j;t),\quad p_f^{(s)}(i,j;t),\quad d_f^{(s)}(i,j;t),
\]
\noindent we collect all nonzero values into three sets for each field \(f\) at time \(t\):

\[
\begin{aligned}
G_{f,t} &: \text{all masked ground truth values}, \\
P_{f,t} &: \text{all masked prediction values}, \\
D_{f,t} &: \text{all masked absolute differences}.
\end{aligned}
\]

We then write
\[
\mu_{\mathrm{gt},f}(t) = \mathrm{mean}\bigl(G_{f,t}\bigr), \quad
\sigma_{\mathrm{gt},f}(t) = \mathrm{std}\bigl(G_{f,t}\bigr),
\]
\[
\mu_{\mathrm{p},f}(t) = \mathrm{mean}\bigl(P_{f,t}\bigr), \quad
\sigma_{\mathrm{p},f}(t) = \mathrm{std}\bigl(P_{f,t}\bigr),
\]
\[
\mu_{\mathrm{diff},f}(t) = \mathrm{mean}\bigl(D_{f,t}\bigr), \quad
\sigma_{\mathrm{diff},f}(t) = \mathrm{std}\bigl(D_{f,t}\bigr).
\]

This way, we avoid cumbersome index lists while still capturing the full aggregation over samples and spatial locations.

\subsubsection{Sample-wise Metrics}

\noindent\emph{(Here and below, we fix $\ell$ to its final autoregressive value and omit it from the notation for brevity.)}

For each sample \(s\), let
\[
\begin{aligned}
M^{(s)}(t) = \bigl\{(i,j)\mid M_{\mathrm{g}}(s,t,\ell,i,j)=1\bigr\}, \\
M^{(s)}_{\mathrm{p}}(t) = \bigl\{(i,j)\mid M_{\mathrm{p}}(s,t,\ell,i,j)=1\bigr\}.
\end{aligned}
\]

\noindent\textit{Mask convention.} All masked quantities are evaluated on the ground truth material region $M^{(s)}(t)$, so predicted and ground truth fields are compared on the same set of grid points.

\begin{itemize}
  \item \textbf{Root mean square error (RMSE)}
    \begin{equation}
    \begin{aligned}
    \mathrm{RMSE}_f^{(s)}(t)
    &= \Biggl[
    \frac{1}{\lvert M^{(s)}(t)\rvert}
    \\
    &\qquad \sum_{(i,j)\in M^{(s)}(t)}
    \bigl(p_f^{(s)}(i,j;t) - g_f^{(s)}(i,j;t)\bigr)^2
    \Biggr]^{1/2}.
    \end{aligned}
    \end{equation}

  \item \textbf{Coefficient of determination (\(R^2\))}
  \begin{equation}
  R^{2,(s)}_{f}(t)
  = 1 \;-\;
    \frac
      {\displaystyle\sum_{(i,j)\in M^{(s)}(t)}
        \bigl(p_f^{(s)}(i,j;t)-g_f^{(s)}(i,j;t)\bigr)^{2}}
      {\displaystyle\sum_{(i,j)\in M^{(s)}(t)}
        \bigl(g_f^{(s)}(i,j;t)-\bar g_f^{(s)}(t)\bigr)^{2}},
  \end{equation}
  where
  \[
    \bar g_f^{(s)}(t)
    = \frac{1}{\lvert M^{(s)}(t)\rvert}
      \sum_{(i,j)\in M^{(s)}(t)} g_f^{(s)}(i,j;t).
  \]
  
  \item \textbf{Intersection over union (IoU)}
  \begin{equation}
  \mathrm{IoU}_f^{(s)}(t)
  = 
  \frac
    {\lvert M^{(s)}(t)\cap M^{(s)}_{\mathrm{p}}(t)\rvert}
    {\lvert M^{(s)}(t)\cup M^{(s)}_{\mathrm{p}}(t)\rvert}.
  \end{equation}
\end{itemize}

We then aggregate across samples $s \in \{1,\dots,N\}$ to compute overall means and standard deviations as before.

\subsubsection{Field Moments}

For each sample \(s\), the masked mean and standard deviation of field \(f\) at time \(t\) are
\[
\mu_{\mathrm{g},f}^{(s)}(t)
= 
\frac{1}{\lvert M^{(s)}(t)\rvert}
\sum_{(i,j)\in M^{(s)}(t)}
  g_f^{(s)}(i,j;t),
\]
\[
\sigma_{\mathrm{g},f}^{(s)}(t)
= 
\mathrm{std}\!\bigl(\{\,g_f^{(s)}(i,j;t)\mid(i,j)\in M^{(s)}(t)\}\bigr),
\]
\[
\mu_{\mathrm{p},f}^{(s)}(t)
= 
\frac{1}{\lvert M^{(s)}(t)\rvert}
\sum_{(i,j)\in M^{(s)}(t)}
  p_f^{(s)}(i,j;t),
\]
\[
\sigma_{\mathrm{p},f}^{(s)}(t)
= 
\mathrm{std}\!\bigl(\{\,p_f^{(s)}(i,j;t)\mid(i,j)\in M^{(s)}(t)\}\bigr).
\]

\end{document}